%% file: main.tex
\documentclass[10pt,twocolumn,letterpaper]{article}

\usepackage[pagenumbers]{cvpr} 
\usepackage[pagebackref,breaklinks,colorlinks,citecolor=cvprblue]{hyperref}
\usepackage{booktabs}
\usepackage{multirow}
\usepackage{balance}
\usepackage[dvipsnames]{xcolor}
\usepackage{array}
\definecolor{cvprblue}{rgb}{0.21,0.49,0.74}
\definecolor{myblue}{RGB}{46, 105, 167}
\definecolor{myorange}{RGB}{186, 82, 30}
\definecolor{ForestGreen}{rgb}{0.13, 0.55, 0.13}
\definecolor{Maroon}{rgb}{0.69, 0.19, 0.0}
\definecolor{my_pink}{rgb}{1,0.43,0.41}
\definecolor{my_gray}{rgb}{0,0,0.2}
\definecolor{citecolor}{RGB}{30,130,255}
\definecolor{LightCyan}{rgb}{0.88,1,1}

\newcommand\mypara[1]{\vspace{1mm}\noindent\textbf{#1}}

\def\paperID{13238} 
\def\confName{CVPR}
\def\confYear{2025}

\title{ Raccoon: Multi-stage Diffusion Training with Coarse-to-Fine Curating Videos}

\author{
Zhiyu Tan $^{14}$ \quad
Junyan Wang $^{2}$ \quad
Hao Yang $^{3}$ \quad
Luozheng Qin $^{3}$ \quad 
Hesen Chen $^{4}$ \\ [2pt]\quad 
Qiang Zhou $^{3}$  \quad
Hao Li $^{14}$\thanks{Corresponding author}\\ [6pt]
{
$^{1}$ {Fudan University} \quad $^{2}$ {The University of Adelaide} \quad $^{3}$ {INF Tech}
} \\
{$^{4}$ {Shanghai Academy of Artificial Intelligence for Science}}
} 

\newcommand{\ourmethod}{{\textbf{\textsc {Raccoon}}}\xspace}
\newcommand{\ourmethodbold}{{\textbf{\textsc {Raccoon}}}\xspace}
\newcommand{\ourdataset}{{\textbf{\textsc {CFC-Vids-1M}}}\xspace}

\begin{document}

\let\oldtwocolumn\twocolumn
\renewcommand\twocolumn[1][]{%
    \oldtwocolumn[{#1}{
    \vspace{-3em}
    \begin{center}
           \includegraphics[width=0.9\textwidth]{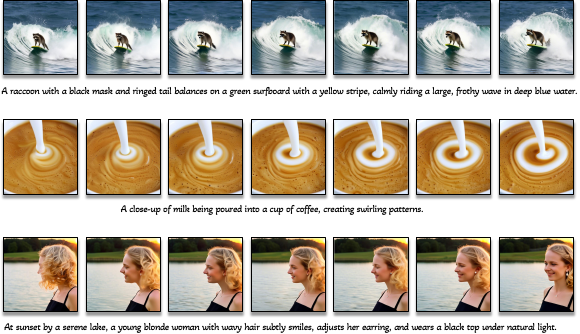}
           \captionof{figure}{\ourmethodbold Samples for Text-to-Video Generation.
            Our approach can generate high-resolution, temporally consistent, photorealistic videos from text prompts. The samples displayed have a resolution of $512 \times 512$, last 4 seconds, and play at 8 frames per second.}
           \label{fig:poster}
        \end{center}
    }]
}
\maketitle

\input{sec/0_abstract}  
\begin{table*}[t]
    \centering
    \small
    \caption{\textbf{Comparison of our dataset and other popular video-text datasets.} We've compiled statistics on video-text datasets, specifically analyzing metrics such as the domain, the number of video clips after scene detection, average duration, average text length.} 
    \label{tab:dataset}
    \begin{tabular}{llccccccc}
    \toprule
    Dataset & Venue~\&~Year & Text & Domain & Clip Num & Avg Video Len & Avg Text Len & Resolution \\
    \midrule
    MSVD~\citep{chen2011msvd}               & ACL ~~~2011 & Human    & Open    & 2K   & 9.7s  & 8.7 words  & -     \\
    UCF101~\citep{soomro2012ucf101}         & ICCV ~2013 & Human   & Action   & 13K    & 7.2s     & 4.3 words & 240p \\
    LSMDC~\citep{rohrbach2015lsmdc}                     & CVPR 2015 & Human    & Movie   & 118K   & 4.8s  & 7.0 words  & 1080p \\    ActivityNet~\citep{caba2015activitynet}         & CVPR 2015 & Human    & Action  & 100K   & 36.0s & 13.5 words & -     \\
    MSR-VTT~\citep{xu2016msrvtt}            & CVPR 2016 & Human    & Open    & 10K    & 15.0s & 9.3 words  & 240p  \\
    DiDeMo~\citep{anne2017ldidemo}          & ICCV ~2017 & Human    & Flickr  & 27K    & 6.9s  & 8.0 words  & -     \\
    YouCook2~\citep{zhou2018youcook2}               & AAAI ~2018 & Human    & Cooking & 14K    & 19.6s & 8.8 words  & -     \\
    VATEX~\citep{wang2019vatex}             & ICCV ~2019 & Human    & Open    & 41K    & 10s & 15.2 words & - \\
    HowTo100M~\citep{miech2019howto100m}    & ICCV ~2019 & ASR               & Open    & 136M   & 3.6s  & 4.0 words  & 240p  \\
    YT-Temporal-180M~\citep{zellers2021yt180m} & NIPS ~~2021 & ASR            & Open    & 180M   & -     & -          & -     \\
    HD-VILA-100M~\citep{xue2022hdvila}     & CVPR 2022 & ASR                & Open    & 103M   & 13.4s & 32.5 words & 720p  \\
    Panda-70M~\citep{chen2024panda}                             & CVPR 2024 & Auto & Open    & 70.8M  & 8.5s  & 13.2 words & 720p \\
    \midrule
    \textbf{\ourdataset }                          &   & Auto & Open    & 1M     & 10.6s & 89.3 words          & 720p \\
    \bottomrule
    \end{tabular}
    \vspace{-3.5mm}
\end{table*}

\input{sec/1_intro}

\input{sec/2_related_work}
\input{sec/3_method_dataset}

\input{sec/3_method_training}
\input{sec/4_experiment}

\input{sec/5_conclusion}
{
    \small
    \bibliographystyle{ieeenat_fullname}
    \bibliography{main}
}
\clearpage \input{sec/6_suppl}

\end{document}

%% file: sec/0_abstract.tex
\begin{abstract}

Text-to-video generation has demonstrated promising progress with the advent of diffusion models, yet existing approaches are limited by dataset quality and computational resources.
To address these limitations, this paper presents a comprehensive approach that advances both data curation and model design. We introduce \ourdataset, a high-quality video dataset constructed through a systematic coarse-to-fine curation pipeline. The pipeline first evaluates video quality across multiple dimensions, followed by a fine-grained stage that leverages vision-language models to enhance text-video alignment and semantic richness.
Building upon the curated dataset's emphasis on visual quality and temporal coherence, we develop \ourmethodbold, a transformer-based architecture with decoupled spatial-temporal attention mechanisms. The model is trained through a progressive four-stage strategy designed to efficiently handle the complexities of video generation. Extensive experiments demonstrate that our integrated approach of high-quality data curation and efficient training strategy generates visually appealing and temporally coherent videos while maintaining computational efficiency. 
We will release our dataset, code, and models.

\end{abstract}

%% file: sec/1_intro.tex
\section{Introduction}
\label{sec:intro}

Text-to-video generation, which aims to synthesize video sequences from textual descriptions, has emerged as a significant research direction in AIGC. Recent years have witnessed remarkable progress in generative models, particularly in the realm of diffusion-based approaches \cite{brooks2024video}. The success of these models in image synthesis has opened new possibilities for video generation, with potential applications spanning creative content creation, educational resources, and visual storytelling~\cite{gupta2023walt}~\cite{ma2024latte}~\cite{opensora}. The synthesis of  videos that faithfully align with textual descriptions requires both comprehensive training data and sophisticated learning approaches to handle the complexities of spatial-temporal modeling.

Recent text-to-video generation models \cite{ma2024latte,gupta2023walt,blattmann2023align,lu2023vdt,wang2024leo,chen2023seine} have demonstrated promising results by leveraging large-scale video datasets \cite{chen2024panda,xue2022hdvila,miech2019howto100m}. Despite these advances, existing datasets present limitations that affect model training effectiveness. These datasets face two primary challenges: 1) video quality issues, including temporal inconsistencies from imprecise scene detection and prevalence of static content lacking motion dynamics; 2) caption quality limitations, manifested in frame-by-frame descriptions rather than coherent narratives, insufficient descriptive details with limited word counts, and inadequate semantic alignment between text and video content. To address these challenges, we introduce \ourdataset, constructed through a systematic coarse-to-fine curation pipeline that ensures both visual and caption quality.

Our proposed coarse-to-fine curation pipeline addresses dataset quality through two systematic stages. At the coarse level, we evaluate video quality across multiple dimensions including aesthetic appeal, temporal consistency, OCR presence, motion dynamics, and category distribution. Each dimension is quantitatively assessed using specialized models. The fine-grained stage focuses on enhancing text-video alignment through a two-step process that leverages vision-language and large language models, ensuring both caption informativeness and semantic accuracy.

To fully leverage the curated \ourdataset dataset for text-to-video synthesis, efficient architectures and training strategies are crucial. We develop a transformer-based model \ourmethod with decoupled spatial-temporal attention mechanisms, enabling efficient processing of video sequences. Building upon this architecture, we propose a four-stage training strategy that progressively enhances model capabilities: beginning with semantic learning from pre-trained image models, followed by temporal modeling at low resolution, then scaling to high-resolution video generation, and finally refining visual quality through targeted fine-tuning. This systematic approach effectively addresses the computational challenges of video generation while maintaining generation quality.

This paper presents a comprehensive approach to text-to-video generation through both data curation and model design. The proposed coarse-to-fine curation pipeline ensures high-quality training data across multiple dimensions, while the progressive training strategy enables efficient learning of both semantic and temporal aspects. Extensive experiments demonstrate the effectiveness of our approach in generating high-quality and temporally coherent videos.

\begin{figure*}[htb]
    \centering
    \includegraphics[width=\textwidth]{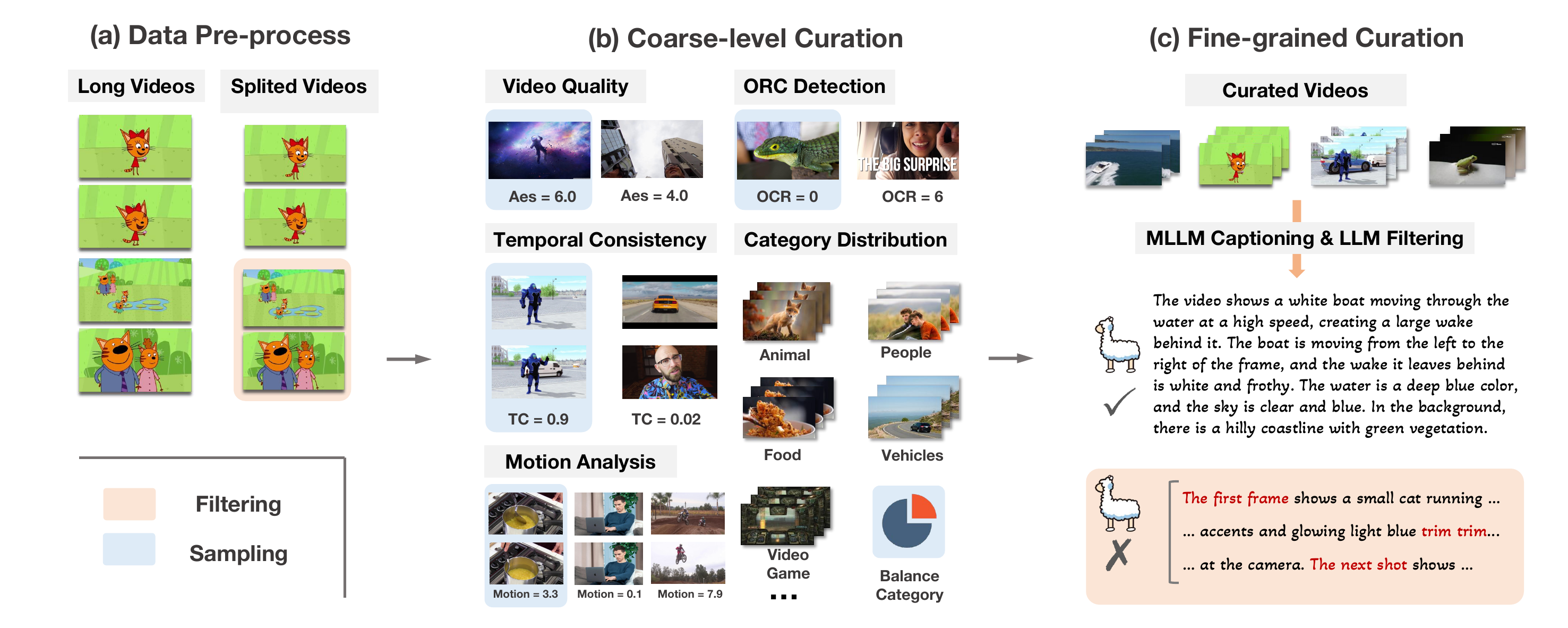}  
    \caption{\textbf{Overview of data curation.}
    Firstly, we employ a scene splitting algorithm to divide long videos with multiple scenes into single scene shots. We filter and sample videos based on five aspects: video quality, OCR, temporal consistency, category, and motion. Finally, we use a Large Language Model (LLM) to curate video-text pairs for error captions. } 
    \label{fig:dataset_overview} 
\end{figure*}

%% file: sec/2_related_work.tex
\section{Related Work}

\mypara{Vision-Language Dataset}.
To advance video understanding and generation, numerous video-text datasets have been developed, differing in aspects such as video length, resolution, domain, and scale. These datasets are compared in Table~\ref{tab:dataset}.
Early datasets like MSVD~\citep{chen2011msvd} and MSRVTT~\citep{xu2016msrvtt} rely on human annotation for video captioning, ensuring high-quality text-video alignment but limiting dataset scale. To address scalability, datasets such as YT-HowTo100M~\cite{miech2019howto100m} and HD-VILA-100M~\cite{xue2022hdvila} adopt automatic annotation through ASR-generated subtitles, though this approach often leads to caption inaccuracies. Recent work Panda-70M~\cite{chen2024panda} utilizes a cross-modality teacher model to generate 70 million text-video pairs, demonstrating the potential of model-based caption generation. In this work, we present a two-stage coarse-to-fine curation pipeline that combines multi-dimensional quality filtering and model-based caption enhancement to ensure both visual quality and text-video alignment.

\mypara{Video Generation Methods}.
Recent advances in video synthesis have explored various architectural paradigms. Autoregressive methods like VQGAN extensions~\cite{hong2022cogvideo,yan2021videogpt} predict sequential video tokens but face computational constraints due to their sequential nature. GAN-based approaches~\cite{jiang2021transgan,lee2021vitgan} achieve efficient generation but often struggle with stability and temporal consistency.
More recently, transformer-based diffusion models have emerged as a promising direction for video generation~\cite{ma2024latte,gupta2023walt,blattmann2023align,lu2023vdt,wang2024leo,chen2023seine,ho2022video,opensora}. These approaches demonstrate superior generation quality and temporal coherence, though at the cost of substantial computational requirements. For instance, W.A.L.T~\cite{gupta2023walt} achieves excellent performance through window-based attention and unified latent space for images and videos. In this work, we present a transformer-based diffusion model trained with a multi-stage strategy for high-quality video generation.

\mypara{Efficient Training Strategies}.
Recent works in diffusion models have explored various approaches to improve training efficiency. Training strategy decomposition methods segment the learning process into sequential stages, as demonstrated in PixelArt~\cite{chenpixart} and OmniDiffusion~\cite{tan2025empirical}, where text-image alignment and visual quality are optimized progressively. Alternatively, model decomposition approaches utilize cascaded sub-models for generation, as implemented in Imagen~\cite{ho2022imagen} and W.A.L.T~\cite{gupta2023walt}.
More recently, these efficient training strategies have been adapted for video generation tasks~\cite{ho2022imagen,gupta2023walt,saharia2022photorealistic}, demonstrating promising results in managing computational constraints while maintaining generation quality. In this work, we propose a progressive multi-stage training strategy that effectively balances computational efficiency and generation quality while maintaining temporal consistency.

\label{sec:related}

%% file: sec/3_method_dataset.tex
\section{Coarse-to-Fine Curation}
\label{sec:method_dataset}

This section describes the construction of \ourmethod, a comprehensively curated video dataset derived from HD-VILA~\cite{xue2022hdvila}. The initial HD-VILA dataset comprises 3.8M high-resolution videos, which are processed into 108M video clips through systematic segmentation. Following established practices~\cite{blattmann2023svd, chen2024panda,opensora}, scene transitions in raw videos are first detected and removed using PySceneDetect~\cite{pyscenedetect} to eliminate motion inconsistencies that could affect model performance. The curation process comprises two main stages: coarse-level filtering for basic quality control and fine-grained refinement for enhanced text-video alignment.

\subsection{Coarse-Level Curation}

\begin{figure*}[t]
    \centering
    \includegraphics[width=\textwidth]{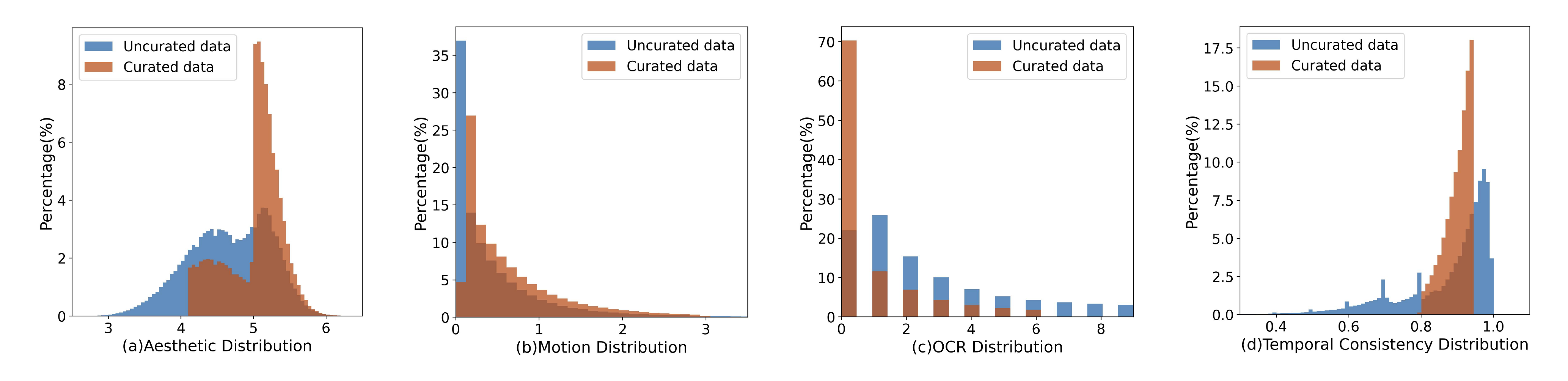}
    \caption{\textbf{Comparison of statistics between uncurated and curated datasets.}
     (a) through (d) present comparative statistics of uncurated and curated datasets across multiple dimensions: (a) aesthetics, (b) motion, (c) Optical Character Recognition (OCR), (d) temporal consistency.
    }
    \label{fig:data_distribution}
\end{figure*}

High-quality training data is crucial for text-to-video synthesis models, requiring careful curation across multiple dimensions including visual quality, category balance, and temporal coherence. To establish baseline quality control, we design a coarse curation strategy that evaluates video segments through five key aspects: video quality assessment, OCR detection, temporal consistency verification, category distribution balancing, and motion analysis. Each video segment is assigned corresponding quality tags to facilitate the filtering process.
Fig \ref{fig:data_distribution} illustrates the comparison between uncurated and curated datasets.

\mypara{Video Quality}.
High-quality visual content fundamentally determines the generation capability of text-to-video models, as the model learns to synthesize videos by replicating the visual characteristics of training data. We employ the LAION Aesthetics model~\cite{schuhmann2022laion} to evaluate and filter videos based on aesthetic scores, eliminating those with low visual appeal. This approach effectively removes visually inappropriate content, including videos with irregular color distributions or undesirable visual elements.

\mypara{OCR Detection}.
The presence of subtitles and text overlays in videos can negatively impact the visual quality of generated content and introduce undesirable patterns in text-to-video synthesis. To address this issue, we employ PP-OCR~\cite {li2022pp_ocr} for automated text detection in video frames. Specifically, we compute OCR scores for keyframes in each video clip and use these scores as filtering criteria during the data curation process. Videos with high OCR scores, indicating substantial text content, are filtered out to maintain the visual purity of our training dataset.

\mypara{Temporal Consistency}.
Incorrect scene splitting in videos can impair model training by introducing semantic gaps in visual flow and content coherence. To address this, we leverage the CLIP model to ensure temporal coherence within video clips. Specifically, we compute the cosine similarity between the initial and final frames to assess frame-level consistency quantitatively. Videos exhibiting low similarity scores, which often indicate sudden scene changes or semantic inconsistencies, are filtered out from the dataset.

\mypara{Motion Analysis}.
The quality of motion representation in training videos directly impacts a model's ability to generate natural and smooth video sequences. To quantify motion characteristics, we utilize the RAFT~\citep{teed2020raft} model to compute optical flow scores for each video clip. Videos exhibiting either minimal motion or excessive movement are filtered out, as both extremes can lead to degraded motion modeling and unrealistic video generation.

\mypara{Category Distribution}.
Category imbalance in training data can substantially degrade model performance, leading to biased generation results and limited diversity across different video types. To address the category imbalance inherent in the existing datasets, we implement a CLIP-based categorization system. Our approach computes the average CLIP features from the initial, middle, and final frames of each video, and then assigns tags based on their similarity to predefined category embeddings.

\begin{figure}
\centering
    \includegraphics[width=.9\linewidth]{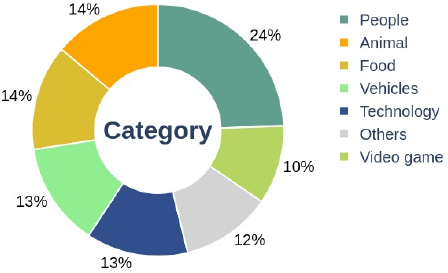}
    \caption{\textbf{The distribution of categories in \ourmethod.} The dataset contains a total of 14 categories, with a balanced distribution across the primary categories.}
    \label{fig:category}
\end{figure}

Based on the established quality assessment, we implement a hierarchical sampling strategy that first filters out videos with quality scores below predefined thresholds. Specifically, videos are removed if they exhibit low aesthetic scores, high OCR presence, poor temporal consistency, and extreme motion patterns. The remaining videos are then sampled according to their category distribution to ensure a balanced representation. Fig.\ref{fig:data_distribution} shows the comparison of statistics between uncurated and curated datasets. These coarse-level quality tags establish the foundation for subsequent fine-grained curation while maintaining diversity across categories.

\begin{table}[t]
\centering
\caption{Statistics of noun and verb concepts for different datasets. 
VN: valid distinct nouns (appearing more than 10 times); DN: total distinct nouns; Avg N: average noun count per video.
VV: valid distinct verbs (appearing more than 10 times); DV: total distinct verbs; Avg N: average verbs count per video.}
\begin{tabular}{l|cccc}
\toprule
\textbf{Dataset}      & \textbf{VN/DN}    & \textbf{VV/DV}    & \textbf{Avg N} & \textbf{Avg V} \\
\midrule
Pandas-70M            & 16.1\%            & 19.2\%            &   4.3             &   1.9      \\
\textbf{Ours}                  & 20.3\%            & 41.1\%            &   22.5             &   15.9     \\
\bottomrule
\end{tabular}

\label{tab:dataset_summary}
\end{table}

\subsection{Fine-Grained Curation}

High-quality text-video alignment is essential for effective text-to-video synthesis models. Following the coarse-level filtering, a two-step fine-grained curation strategy is proposed that focuses on generating informative captions and filtering problematic text-video pairs.

Despite CLIP-based filtering in coarse-level curation, several critical issues persist in video captioning: scene transition errors where captions fail to recognize scene boundaries, token generation failures resulting in repetitive content, and frame-level descriptions that lack temporal coherence. 
To address these challenges, a two-phase curation is implemented. First, the state-of-the-art vision-language model ViLA~\cite{lin2024vila} is employed to generate informative and detailed video captions. Subsequently, LLAMA~\cite{dubey2024llama} is utilized to identify and filter out problematic captions through prompt-based evaluation:
\begin{quotation}
\label{quote:llama-prompt}
\textit{Please respond with `Yes' or `No' to the following questions:}
\begin{itemize}
   \item \textit{Given the preceding video caption, is there an indication of a possible scene transition?}
   \item \textit{Does the preceding video caption suggest a shift towards a series of descriptive image captions?}
   \item \textit{Does the video caption conclude with repetitive phrases or sentences? }
\end{itemize}
\end{quotation}

As quantitatively demonstrated in Table \ref{tab:dataset_summary}, the enhanced captions show substantial improvements in vocabulary diversity and semantic density. Through this systematic fine-grained curation approach, both caption quality and text-video alignment are significantly enhanced, enabling more precise control over video generation through enriched textual descriptions.

%% file: sec/3_method_training.tex
\section{\ourmethod}
\label{sec:method_training}

Our method consists of a transformer-based diffusion architecture and a systematic training strategy that enables efficient text-to-video synthesis. The model architecture leverages efficient latent space processing, while the training strategy progressively enhances generation capabilities across multiple stages.

\begin{figure*}
\centering
    \includegraphics[width=0.9\textwidth]{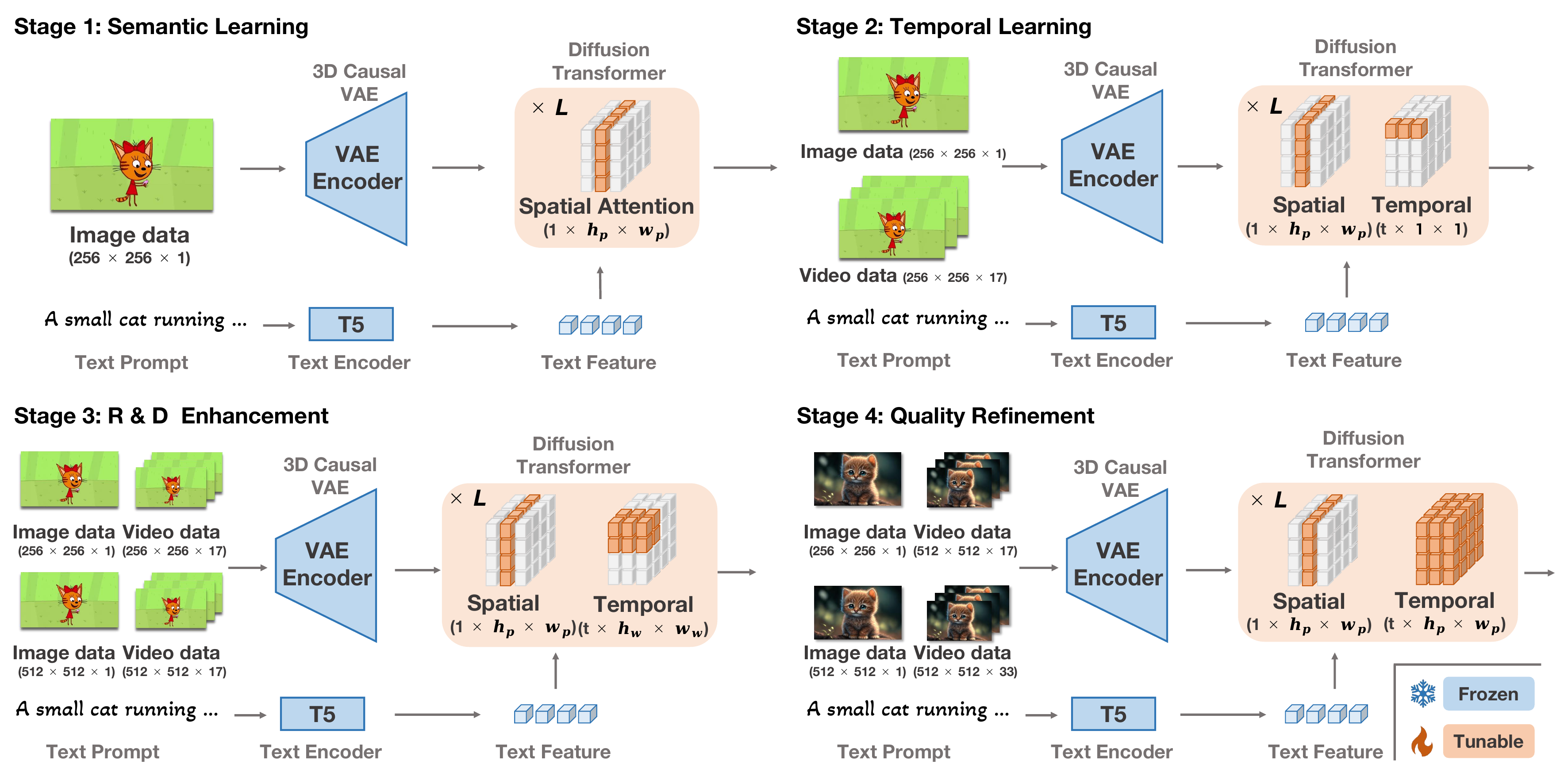}
    \caption{\textbf{Four-stage training pipeline.} 
    Leverages pre-trained text-to-image models to establish semantic understanding capabilities as the foundation for video generation.
    Jointly trains image and video data at low resolution to efficiently optimize temporal modules.
    Enhances spatial details and temporal coherence through high-resolution training, enabling long video generation.
    Fine-tunes the model using a curated high-quality dataset to improve visual consistency and aesthetic quality of generated videos.}
    \label{fig:training_overview}
\end{figure*}

\subsection{Model Architecture}
Given an input video $x \in \mathbb{R}^{(1+T) \times H \times W \times C}$, we employ a 3D Causal VAE to address the computational challenges in video generation through efficient dimensionality reduction. The encoder maps the input into a low-dimensional representation $z \in \mathbb{R}^{(1+t) \times h \times w \times c}$, achieving compression along both spatial ($f_s = H/h = W/w$) and temporal ($f_t = T/t$) dimensions. By independently encoding the first frame, this architecture establishes a unified latent space that effectively bridges image and video domains.

The transformer backbone processes these latent representations through a series of attention blocks. The input latents are first patchified with size $p$ and enhanced with positional information through a combination of 2D spatial and 1D temporal embeddings based on sinusoidal functions~\cite{vaswani2017attention}.

\subsection{Training Strategy}

An efficient text-to-video generation model requires a well-structured training strategy capable of handling the complex nature of synthesizing video content across both spatial and temporal dimensions. Thus, we present a four-stage training pipeline that progressively enhances the model's capacity, transitioning seamlessly from basic semantic comprehension to sophisticated video generation.
Throughout all stages, the spatial attention mechanism operates within each frame, using tokens of dimensions $1 \times h_p \times w_p$, where $h_p=h/p$, $w_p=w/p$ and $p$ is patchify size.

Simultaneously, the temporal attention mechanism is marked by a systematic expansion of the 3D window size. Beginning from an initial size of $(t \times 1 \times 1)$, the window progressively accommodates the total number of tokens in the video latent size $(t \times h_p \times w_p)$. This adaptive approach facilitates the model in incrementally learning and synthesizing more complex spatio-temporal patterns. A comprehensive illustration of this four-stage training can be found in Fig.\ref{fig:training_overview}.

\mypara{Stage 1: Semantic Learning}.
Text-to-video generation requires robust semantic understanding while training such capabilities directly on limited video datasets remains computationally intensive and inefficient. To address this limitation, we leverage pre-trained text-to-image models that have developed comprehensive semantic knowledge from large-scale image datasets. 
Specifically, we employ a pre-trained Causal 3D VAE to establish a unified latent space for both images and videos. This shared representation space allows us to effectively bridge image and video domains by treating images as single-frame videos. During this stage, we exclusively optimize the semantic module parameters with through text-to-image tasks. This unified approach enables efficient knowledge transfer from image pre-training, allowing subsequent stages to focus on temporal dynamics.

\mypara{Stage 2: Temporal Learning}.
The optimization process of temporal modules necessitates significant computational resources due to its complexity. To improve training efficiency, we propose to jointly train the model with low-resolution data.
Specifically, we conduct joint training with both image-text and video-text pairs at $256 \times 256$ resolution. This joint optimization strategy enables fast convergence of temporal modules while maintaining semantic consistency through continued image-text training. It's important to note that a temporal attention size of $(t \times 1 \times 1)$ is employed for efficient training. The low-resolution training approach effectively balances computational efficiency and temporal modeling capability, establishing the foundation for subsequent training stages.

\mypara{Stage 3: Resolution \& Duration Enhancement}.
While Stage 2 establishes temporal modeling capabilities, the low-resolution outputs remain insufficient for practical applications that demand high-quality video generation. To address this limitation, we extend our training to high-resolution, long-duration video synthesis.
Specifically, we fine-tune the model using high-resolution images and videos while maintaining the joint training strategy from Stage 2. In contrast to Stage 2 training, we apply a temporal attention size of $(t \times h_w \times w_w)$ to enhance temporal consistency. Here, $h_w$ and $w_w$ denote the height and width of the 3D sub-window size among video latents. Specifically, we set both $h_w$ and $w_w$ to 8.  This approach enhances both spatial details and temporal coherence over extended durations, enabling the model to generate videos that meet contemporary quality standards. The transition to high-resolution training improves the model's capability to produce visually detailed and temporally consistent videos.

\mypara{Stage 4: Quality Refinement}
Despite achieving high-resolution video generation in Stage 3, the output quality exhibits inconsistency in visual aesthetics. To stabilize generation quality and enhance visual appeal, we implement a focused fine-tuning strategy.
Specifically, from our curated dataset, we further select 50K videos and 100k images with stricter quality criteria, requiring aesthetic scores above 5.5 for video and 7.0 for image. This carefully filtered dataset, supplemented by human evaluation, guides the model toward generating visually consistent and refined videos. The fine-tuning process stabilizes generation quality while enhancing the visual aesthetics of the generated videos. During this training stage, we use 3D full temporal attention, represented as $t \times h_p \times w_p$, to further improve the quality of the generated videos.

Through this progressive training pipeline, our model evolves from basic semantic understanding to high-quality video generation capabilities. Each stage addresses specific challenges, from semantic learning to aesthetic refinement, enabling our model to generate visually refined and temporally coherent videos from text prompts.

%% file: sec/4_experiment.tex
\section{Experiment}
\label{sec:experiment}

\mypara{Datasets.}
For the ablation study of the training strategy, we utilize the standard video benchmark UCF-101~\cite{soomro2012ucf101} for class-conditional generation. All training and testing splits from UCF-101 are used for training purposes. For the text-to-video generation, we employ a joint training approach on text-image and text-video pairs. The training dataset consists of 43 million text-image pairs and three categories of text-video pairs: (1) a complete version with 4 million pairs, (2) an open-source version with 1 million curated pairs, and (3) an open-source version with 1.5 million uncurated pairs.

\mypara{Evaluation metrics.}
For quantitative comparisons, we utilize three evaluation metrics: Fréchet Video Distance (FVD)~\cite{unterthiner2018towards_fvd}. Our primary emphasis is on FVD, given that its image-based counterpart, FID, closely corresponds with human subjective judgment. In accordance with the evaluation guidelines established by StyleGAN-V~\cite{skorokhodov2022stylegan}, FVD scores are computed by analyzing 2,048 video clips in zero-shot manner.
To further investigate the visual quality of video produced by \ourmethod, we conduct a user study to discern human preferences among different models.
In this study, we ask users to rate the videos generated by each model on a scale of 1 to 5 across five dimensions: faithfulness, text-video alignment, temporal consistency, dynamic degree, and imaging quality. For more details, please refer to the supplementary materials.

\mypara{Implementation details.}
We employ the AdamW optimizer for training all models, using a constant learning rate of $2 \times 10^{-4}$. The only data augmentation technique we utilize is horizontal flipping. In accordance with standard practices in generative modeling studies~\cite{peebles2023dit}, we maintain an Exponential Moving Average (EMA) of Latte weights throughout the training process, using a decay rate of 0.9999. All the results reported in this study are directly obtained from the EMA model. 

\subsection{Ablation Study}
~\label{ablation_exp}
To validate the effectiveness of each stage in our four-stage training pipeline, we conduct ablation studies by removing different stages while keeping other components unchanged. The quantitative results are presented in Tab. ~\ref{tab:ablation_training_strategy}. 
By prioritizing the training of semantic modules early in Stage 1, computational costs for subsequent stages are reduced, facilitating faster convergence in Stage 2. The integration of Stage 1 and Stage 2 training strengthens the semantic foundation and accelerates the training of temporal modules. This efficient progression is important for improving training efficiency in Stage 3, which focuses on processing high-resolution, long-duration video content, leading to substantially faster convergence.
Note that, Stage 4 is optimized for high aesthetic quality generation. Therefore, we have not conducted experiments on Stage 4 within the UCF101 dataset.

\begin{table}[t]
\centering
\small
\caption{\small Ablative results of different training stages. 
                Under the same computational resources, Stage 1 training efficiently primes the model's semantic modules, significantly accelerating the convergence of subsequent stages, particularly Stage 2 for temporal modules, and culminating in faster training of high-resolution, long-duration videos in Stage 3.}
\vspace{-2mm}
\resizebox{\linewidth}{!}{
\begin{tabular}{cccccc} \toprule
\multicolumn{3}{c}{Training Stages} & \multirow{2}{*}{Res \& Frames} & \multirow{2}{*}{FVD} & \multicolumn{1}{c}{GPU} \\ 
 stage 1 & stage 2 & stage 3    & \multicolumn{1}{c}{} & \multicolumn{1}{c}{}   & \multicolumn{1}{c}{Days}   \\ 
\midrule
            & \checkmark &                     & $256 \times 256 \times 17$ & 267 &   1.5     \\
 \checkmark & \checkmark &                     & $256 \times 256 \times 17$ & 144 &   1.5     \\
            &            & \checkmark          & $512 \times 512 \times 33$ & 478   &    3     \\
 \checkmark & \checkmark & \checkmark          & $512 \times 512 \times 33$ & 313   &    3     \\

\bottomrule
\end{tabular}
}
\label{tab:ablation_training_strategy}
\end{table}

\subsection{Results}

\mypara{Quantitative Results}
\label{qunt_exp}
We evaluate our method against state-of-the-art text-to-video generation techniques using the FVD metric on the UCF-101 dataset, considering various resolutions and durations. Our 1B text-to-video model is trained on a diverse datasets: a 1M curated dataset, a 1.5M uncurated dataset, and a 4M curated dataset. As shown in Table~\ref{tab:sota_comp}, the results demonstrate that the model, trained using our proposed curation pipeline, consistently outperforms others across multiple resolutions and durations. 
Additionally, our 3B video generation model, trained on the 4M curated dataset, also achieves the best performance. For a fair comparison, we standardize the sampling steps for both Latte~\cite{ma2024latte} and Opensora1.2~\cite{opensora} to 100.

\begin{table}[t]
\centering
\caption{Comparison of Models on UCF101, where the symbol $\ast$ indicates evaluation based on 10k generated videos, while others are based on 2048 videos. $\S$ denotes uncleaned data.}
\resizebox{\columnwidth}{!}{%
\begin{tabular}{lcccc}
\toprule
      \multirow{2}{*}{Method} &
      \multirow{2}{*}{Scale} &
      \multirow{1}{*}{(\#) P-T } &
      \multirow{2}{*}{Res \& Frames} &
      \multicolumn{1}{c}{UCF101}      \\
      \multicolumn{1}{c}{} & \multicolumn{1}{c}{} & \multicolumn{1}{c}{Videos}   & \multicolumn{1}{c}{}   &
      \multicolumn{1}{c}{FVD}   \\
\midrule

CogVideo~\cite{hong2022cogvideo}     &  15.5B   & 5.4M & $256 \times 256 \times 16$ & 701 \\ 
MagicVideo~\cite{zhou2022magicvideo} &  -       & 10M  & $256 \times 256 \times 16$ & 699 \\
Make-A-Video~\cite{singer2022make-a-video} & 9.7B & -  & $256 \times 256 \times 16$ & 367 \\ 
PYoCo~\cite{ge2023pyoco} & 0.3B & - & $256 \times 256 \times 16$ & 355\\ 
LVDM~\cite{he2022lvdm} & 1.2B & 18K & $256 \times 256 \times 16$ & 641 \\ 
ModelScope~\cite{wang2023modelscope}& 1.7B & 10M & $256 \times 256 \times 16$ & 639 \\ 
VideoLDM~\cite{blattmann2023videoldm} & 4.2B & 10M & $256 \times 256 \times 16$ & 550 \\
AnimateDiff$^{\ast}$~\cite{guo2023animatediff} & - & 2M & $256 \times 256 \times 16$ & 421 \\ 
ours & 1B & 1M & $256 \times 256 \times 17$ & 387  \\
\midrule
Latte~\cite{ma2024latte} & 1B & - & $512 \times 512 \times 16$ & 463 \\
OpenSorav1.2~\cite{opensora} & 1.2B & 10M & $512 \times 512 \times 17$ & 1274 \\
ours  & 1B & 1M & $512 \times 512 \times 17$ & 437 \\
\midrule
OpenSorav1.2~\cite{opensora} & 1.2B & 10M & $512 \times 512 \times 33$ & 1355 \\
ours & 1B & 1M & $512 \times 512 \times 33$ &  469 \\
ours & 1B & 1.5M$\S$ & $512 \times 512 \times 33$ & 705 \\
ours & 1B & 4M & $512 \times 512 \times 33$ & 435 \\
ours & 3B & 4M & $512 \times 512 \times 33$ & 412 \\
\bottomrule
\end{tabular}%
}
\label{tab:sota_comp}
\end{table}

\begin{figure*}[t]
    \centering
    \includegraphics[width=\textwidth]{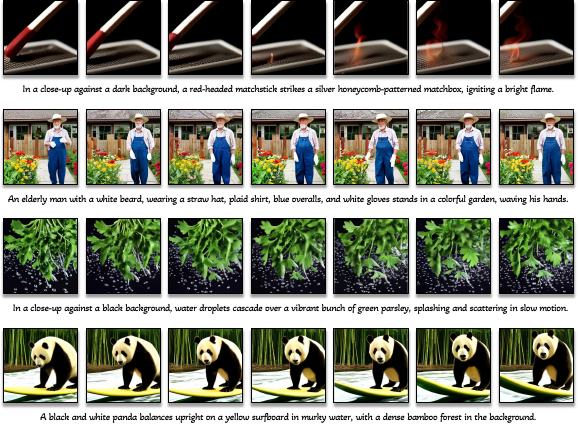}
    \caption{Qualitative Results. Example videos generated by our method at a resolution of 512 × 512 pixels, with a duration of 4 seconds at 8 frames per second. Our model is capable of generating temporally consistent, photorealistic videos that align with the provided prompts.}
    \label{fig:qualitative_results}
\end{figure*}

\mypara{Qualitative Results}
\label{qual_exp}
Figure~\ref{fig:qualitative_results} illustrates the video generation results from \ourmethod based on various prompts.
\ourmethod consistently delivers realistic, high-resolution, and long-duration video generation results across all scenarios. It effectively captures detailed motion and maintains temporal consistency, ensuring that the generated videos align closely with the provided textual prompts. Whether the prompts involve complex actions, subtle movements, or intricate scenes, our model demonstrates its ability to produce photorealistic videos that are both visually compelling and contextually accurate.

%% file: sec/5_conclusion.tex
\section{Conclusion}
In conclusion, this paper introduces a robust methodology that significantly advances data curation and model training strategies for video generation. We present \ourdataset, a meticulously curated, high-quality video dataset developed through a systematic coarse-to-fine pipeline. This process not only enables rigorous multi-dimensional video quality assessment but also employs cutting-edge language models to enhance text-video alignment and enrich captioning detail.
Complementing \ourdataset, we propose \ourmethodbold, a transformer-based model architecture designed with a focus on prompt adherence, visual quality, and temporal consistency. By employing decoupled spatial-temporal attention mechanisms and a progressive four-stage training strategy, our model optimizes computational efficiency and enhances video generation fidelity.
Our extensive experimental validation demonstrates that \ourmethodbold consistently generates visually engaging and temporally coherent videos, all while maintaining impressive computational efficiency. The pioneering methodologies introduced in this work establish a solid foundation for future research, paving the way for more advanced, high-fidelity video generation models that push the boundaries of current capabilities in this field. We anticipate these contributions will inspire further innovations.
\label{sec:conclusion}

%% file: sec/6_suppl.tex

\appendix
\section{Ethical and Social Impacts}

The proposed text-to-video curation pipeline and efficient training strategy represent a significant leap forward in generative AI capabilities, making video production more accessible, cost-effective, and scalable. These advancements have the potential to empower creators across diverse industries such as education, entertainment, marketing, and accessibility technologies, enabling them to produce high-quality video content with minimal resources.

However, these advancements are accompanied by ethical challenges, particularly the potential misuse of text-to-video technology. One primary concern is the generation of realistic but deceptive content, such as deepfakes, which could be used to propagate misinformation, defame individuals, or manipulate public opinion. Furthermore, the automated curation pipeline might inadvertently propagate harmful stereotypes or biases present in training datasets, raising concerns about unintended discrimination or harm.

On a broader societal level, text-to-video technology could disrupt traditional media production industries, potentially displacing jobs in areas such as video editing, animation, and scriptwriting. Moreover, the increased availability of hyper-realistic AI-generated content may blur the lines between authentic and synthetic media, challenging the public’s ability to discern credible sources. Additionally, access to such technologies could exacerbate existing digital divides if it remains available only to resource-rich organizations.

To address these challenges, we propose integrating transparency mechanisms, such as metadata tagging to clearly identify AI-generated content, and robust dataset curation to mitigate bias and harmful outputs. Collaboration with policymakers and industry stakeholders is crucial to establish ethical guidelines and regulatory frameworks. Furthermore, access to training resources and educational materials could ensure equitable opportunities for all users, minimizing potential digital divides.

Generative AI technologies demand continuous ethical oversight. As text-to-video models evolve, it is imperative to regularly evaluate their societal impact and refine safeguards to prevent misuse. This responsibility extends beyond researchers to involve multidisciplinary efforts, fostering an ecosystem where generative AI benefits society while minimizing harm.

\section{Limitations}

While our proposed approach demonstrates significant advancements in text-to-video generation, several limitations persist. Addressing these challenges presents key opportunities for future research and development. Below, we outline the main limitations and potential solutions:

    \mypara{Computational Resources}.
    Despite the efficiency of our training strategy, large-scale text-to-video models still require substantial computational resources. This constraint may restrict the accessibility of such technologies to resource-rich organizations, especially during the initial training phases. Future work should explore further optimizations to make these models more accessible in resource-constrained environments.

\mypara{Dataset Quality and Diversity}.
    The performance of our model is closely tied to the quality and diversity of the training dataset. While significant effort was made to curate a representative dataset, certain domains or cultural nuances may remain underrepresented, limiting the model's ability to generate contextually accurate or culturally sensitive videos. Expanding and refining the dataset to cover a broader range of contexts is a critical area for improvement.

\mypara{Temporal Consistency and High-Resolution Generation}.
    The model occasionally struggles to maintain temporal consistency in complex or fast-paced video sequences. Additionally, generating high-resolution videos at scale presents challenges due to the trade-off between computational efficiency and output quality. These limitations highlight the need for advanced temporal modeling techniques and optimization strategies to enhance video resolution without compromising performance.

\mypara{Semantic Accuracy}.
    The model's performance with ambiguous or abstract text prompts is inconsistent, resulting in videos that may lack semantic precision. Improving the text-video alignment mechanisms and developing more robust embeddings could help address this issue.

\mypara{Model Interpretability}.
    The black-box nature of deep generative models limits understanding of how specific features or patterns influence generated video content. Developing interpretable architectures or incorporating explainability into the pipeline remains an open area of exploration.

\mypara{Ethical and Regulatory Considerations}.
    Although safeguards were proposed to mitigate misuse, these measures are not foolproof and depend on downstream enforcement mechanisms. The rapidly evolving landscape of ethical and regulatory standards for generative AI will require periodic updates to the pipeline to ensure compliance and responsible use.


\section{\ourdataset Details}
\subsection{Coarse-level Curation Details}

\mypara{Video Quality}.
    To assess the quality of a video clip, we sample three frames: the start frame ($f_{\text{start}}$), the middle frame ($f_{\text{mid}}$), and the end frame ($f_{\text{end}}$). For each frame, an aesthetic score is computed based on the LAION Aesthetic v2 model~\cite{schuhmann2022laion}, denoted as $S^a_{\text{start}}$, $S^a_{\text{mid}}$, and $S^a_{\text{end}}$. The overall Video Quality Score is calculated as the average of these three scores:
    \[
    S_{\text{quality}} = \frac{S^a_{\text{start}} + S^a_{\text{mid}} + S^a_{\text{end}}}{3}.
    \]

\mypara{OCR Score}.
    To evaluate the presence of textual content, we sample three frames: the start frame ($f_{\text{start}}$), the middle frame ($f_{\text{mid}}$), and the end frame ($f_{\text{end}}$). An OCR detection model~\cite{li2022pp_ocr} is used to count the number of detected text regions in each frame, denoted as $S^c_{\text{start}}$, $S^c_{\text{mid}}$, and $S^c_{\text{end}}$. The final \textbf{OCR Score} is the average of these counts:
    \[
    S_{\text{ocr}} = \frac{S^c_{\text{start}} + S^c_{\text{mid}} + S^c_{\text{end}}}{3}.
    \]

\mypara{Temporal Consistency}.
    Temporal consistency measures the similarity of visual features between the start frame ($f_{\text{start}}$) and the end frame ($f_{\text{end}}$). Visual features are extracted using the CLIP image encoder~\cite{radford2021clip}, denoted as $\phi(f_{\text{start}})$ and $\phi(f_{\text{end}})$. The \textbf{Temporal Consistency Score} is computed as the cosine similarity between these feature vectors:
    \[
    S_{\text{tc}} = \text{sim}(\phi(f_{\text{start}}), \phi(f_{\text{end}})),
    \]
    where $\text{sim}$ denotes cosine similarity.

\mypara{Motion Score}.
    To assess motion in the video, we sample three frames: the start frame ($f_{\text{start}}$), the middle frame ($f_{\text{mid}}$), and the end frame ($f_{\text{end}}$). An optical flow model~\cite{teed2020raft} is used to compute motion flow scores for the transitions from start to middle ($m_{\text{start} \rightarrow \text{mid}}$) and middle to end ($m_{\text{mid} \rightarrow \text{end}}$). The final \textbf{Motion Score} is the average of these flow scores:
    \[
    S_m = \frac{m_{\text{start} \rightarrow \text{mid}} + m_{\text{mid} \rightarrow \text{end}}}{2}.
    \]

\mypara{Category}.
    Our proposed method addresses the task of zero-shot video classification by leveraging CLIP~\cite{radford2021clip}, a pre-trained vision-language model. The approach consists of three key steps: sampling representative frames from the video, extracting visual features using the CLIP image encoder, and performing similarity-based classification against pre-defined class tags.
    
    Given a video \(V\) and a set of pre-defined class tags \(\{c_1, c_2, \ldots, c_n\}\), the goal is to assign \(V\) to the most semantically relevant class tag. Specifically, the class tags used are: People, Animal, Plants, Architecture, Food, Vehicles, Natural Scenery, Urban landscape, Ocean, Outer space, Video game, 2D cartoon, 3D cartoon, Technology. Unlike conventional methods requiring labeled datasets and extensive training, our approach utilizes CLIP's shared image-text embedding space to perform classification in a zero-shot setting.
    
    \begin{itemize}
        \item \textbf{Frame Sampling:}  
        To represent the video, we sample three key frames: the start frame (\(f_{\text{start}}\)), the middle frame (\(f_{\text{mid}}\)), and the end frame (\(f_{\text{end}}\)). These frames capture the video's temporal structure and provide a concise summary of its content.
        
        \item \textbf{Class Tag Conversion:}  
        Each class tag \(c_i\) is converted into a natural language prompt \(p_i\), such as "a photo of a animal" for the tag "animal." These prompts align with the training paradigm of CLIP's text encoder, which was trained on descriptive textual data.
    
        \item \textbf{Frame Features:}  
        Each sampled frame \(\{f_{\text{start}}, f_{\text{mid}}, f_{\text{end}}\}\) is passed through CLIP's image encoder, generating visual embeddings \(\{\phi(f_{\text{start}}), \phi(f_{\text{mid}}), \phi(f_{\text{end}})\}\), where \(\phi(\cdot)\) represents the CLIP image encoding function.
        
        \item \textbf{Mean Frame Embedding:}  
        To aggregate information across frames, we compute the mean embedding of the three frame features, producing a single feature vector that represents the video:
        \[
        \phi(V) = \frac{\phi(f_{\text{start}}) + \phi(f_{\text{mid}}) + \phi(f_{\text{end}})}{3}.
        \]
    \end{itemize}
    
    To classify the video, the mean video embedding \(\phi(V)\) is compared to text embeddings \(\{e_{t1}, e_{t2}, \ldots, e_{tn}\}\), which are derived by encoding the prompts \(\{p_1, p_2, \ldots, p_n\}\) using CLIP's text encoder. Cosine similarity is used to measure the alignment between the video and each class tag:
    \[
    S_j = \frac{\phi(V) \cdot e_{tj}}{\|\phi(V)\| \|e_{tj}\|}, \quad j \in \{1, \ldots, n\}.
    \]
    The predicted class corresponds to the prompt \(p_k\) with the highest similarity score:
    \[
    k = \arg\max_j S_j.
    \]
    This approach efficiently maps the video to the closest semantic class in the shared image-text embedding space.


\subsection{Fine-Grained Curation Details} 
We propose leveraging large language models (LLMs)~\cite{dubey2024llama} for zero-shot caption classification to address several persistent challenges in video captioning. To mitigate the high computational cost of LLM-based curation, we first employ a CLIP-based filtering mechanism to identify and exclude misaligned text-video pairs before applying LLM curation.
The LLM curation process targets the following key issues:
\begin{itemize}
\item \textbf{Frame-Level Descriptions:} Captions that lack temporal coherence, resulting in fragmented or inconsistent narratives across frames;
\item \textbf{Scene Transition Errors:} Errors during the pre-processing stage of scene splitting can result in incoherent captions and disjointed scene transition sequences in the video; and
\item \textbf{Token Generation Failures:} Repetitive content generation, which detracts from the quality and informativeness of the captions.
\end{itemize}

\begin{figure}[t]
    \centering
    \includegraphics[width=0.95\linewidth]{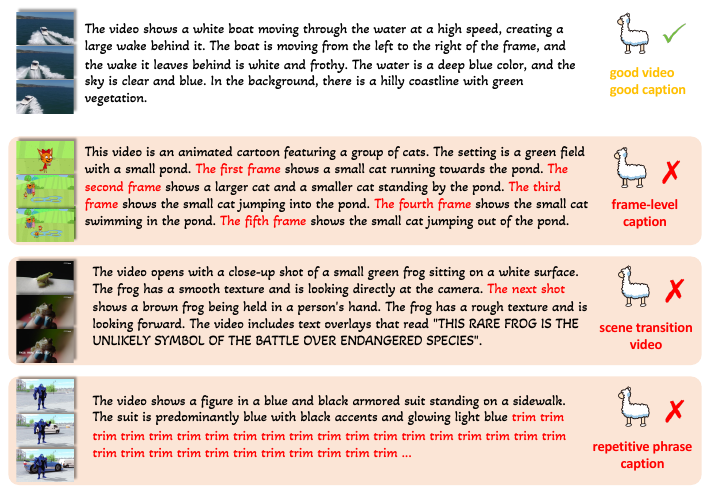} 
    \caption{ \textbf{Examples of text-video alignment using LLM-based zero-shot caption classification.} The figure highlights three common error cases in video captioning: (1) frame-level descriptions that lack temporal coherence; (2) scene transition errors, where captions fail to recognize changes between scenes; and (3) token generation failures, resulting in repetitive content.}
    \label{fig:llm_example}
\end{figure}

As illustrated in Figure~\ref{fig:llm_example}, we present detailed captions for correct text-video pairs alongside examples of three common error cases in text-video alignment.

\section{\ourmethod Details}

\subsection{Background}

\mypara{Diffusion Models}.  
Diffusion models are a class of generative models that learn to produce data through an iterative denoising process. These models begin with samples drawn from a predefined noise distribution and progressively refine them to generate realistic outputs. In the framework of Gaussian diffusion models, the forward noising process gradually adds noise, denoted as $\epsilon$, to real data $(x_0 \sim p_{\text{data}})$. This process is mathematically expressed as:  
\begin{equation}
x_t = \sqrt{\gamma(t)} x_0 + \sqrt{1-\gamma(t)} \epsilon,
\label{eq:add_noise}
\end{equation}
where $t \in [0,1]$, and $\gamma(t)$ is a noise scheduler that monotonically decreases from 1 to 0.  

The reverse process, designed to denoise the corrupted samples, reconstructs clean data by iteratively predicting and subtracting the added noise at each step. The learning objective for this process can be formulated as:  
\begin{equation}
L(\theta) = \mathbb{E}_{\epsilon \sim \mathcal{N}(0, \mathbf{I}), t}\left[ \left\| \epsilon - \epsilon_{\theta}(\mathbf{x}_t; t, c) \right\|^2 \right],
\label{eq:diffusion_loss}
\end{equation}
where $\epsilon_{\theta}$ is the neural network-based denoising model, parameterized by $\theta$. The term $c$ represents input conditions, such as class labels, textual prompts, or other contextual information.  

These properties make diffusion models effective in various generative tasks, including text-to-image and text-to-video synthesis.

\mypara{Latent Diffusion Models (LDMs)}.  
Directly processing high-resolution images and videos using raw pixel representations is computationally intensive. Latent Diffusion Models (LDMs)~\cite{rombach2022high} address this challenge by operating in a compact, lower-dimensional latent space derived from a Vector Quantized-Variational AutoEncoder (VQ-VAE)~\cite{esser2021vqgan}.  

In this framework, a VQ-VAE consists of:  
\begin{itemize}
    \item \textbf{Encoder ($E(x)$)}: Maps an input video $x \in \mathbb{R}^{T \times H \times W \times 3}$ into a latent representation $z \in \mathbb{R}^{t \times h \times w \times c}$, where $T$, $H$, and $W$ denote the temporal, height, and width dimensions of the video, respectively. The downsampling factors are $f_s = H/h = W/w$ for spatial dimensions and $f_t = T/t$ for the temporal dimension.
    \item \textbf{Decoder ($D$)}: Reconstructs the original video $\hat{x}$ from the latent representation $z$.
\end{itemize}

To enhance reconstruction quality, adversarial and perceptual losses---similar to those employed in VQ-GAN---are incorporated into the training process. Operating within the latent space significantly reduces computational costs, making LDMs suitable for generating high-resolution and temporally consistent video data.  

By leveraging LDMs, the proposed text-to-video model achieves efficient synthesis while maintaining high fidelity and scalability.

\subsection{Model \& Training Hyperparameters}
The detailed configurations of the model architecture and the training process, including all relevant hyperparameters, are provided in Table~\ref{tab:hyperparameters1} and Table~\ref{tab:hyperparameters2}. These tables comprehensively outline the parameter settings, enabling precise replication of our experiments and facilitating comparisons with other approaches.

\begin{table}[t]
\centering
\caption{\textbf{Training Hyperparameter Details for Raccoon. }
We show the key training configurations across the four stages of our pipeline. The settings include maximum resolution, duration, batch size, and training steps. Stage 4 (FT) represents the fine-tuning phase.}
\label{tab:hyperparameters1}
\resizebox{\columnwidth}{!}{
\begin{tabular}{lcccc}
\toprule
\textbf{Training Stage} & \textbf{Stage1} & \textbf{Stage2} & \textbf{Stage3} & \textbf{Stage4(FT)} \\ 
\midrule
Max Resolution      & $256 \times 256 $     & $256 \times 256 $    & $512 \times 512 $    & $512 \times 512 $ \\ 
Max duration        &  -       & 2s      & 2s      & 4s   \\ 
Batch Size          & 2048    & 1024   & 1152   & 1024   \\ 
Training Steps      & 100k    & 80k    & 40k    & 10k   \\
\bottomrule
\end{tabular}}
\end{table}

\begin{table}[t]
\centering
\caption{\textbf{Model Hyperparameter Details for Raccoon-1B and Raccoon-3B Models.}
We show the key architectural and training hyperparameters for the Raccoon-1B and Raccoon-3B models. The comparison highlights the scalability of the Raccoon architecture, with Raccoon-3B incorporating larger dimensions and additional capacity to handle more complex tasks.
}
\label{tab:hyperparameters2}
\begin{tabular}{lcc}
\toprule
\textbf{Hyperparameter} & \textbf{Raccoon-1B} & \textbf{Raccoon-3B} \\ 
\midrule
Number of Layers        & 28                 & 32               \\ 
Attention Heads         & 16                 & 24               \\ 
Hidden Size             & 1152               & 1728             \\ 
Position Encoding       & sinusoidal         & sinusoidal       \\ 
Time Embedding Size     & 6912               & 10368            \\ 
Weight Decay            & 0                  & 0                \\ 
Adam $\epsilon$         & 1e-7               & 1e-7             \\ 
Adam $\beta_1$          & 0.9                & 0.9              \\ 
Adam $\beta_2$          & 0.999              & 0.999            \\ 
Learning Rate Decay     & cosine             & cosine           \\ 
Gradient Clipping       & 0.1                & 0.1              \\ 
Text Length             & 200                & 200              \\ 
Training Precision      & bf16               & bf16             \\ 
\bottomrule
\end{tabular}
\end{table}

\section{Additional Results}
\subsection{Human Evaluation}

\begin{figure}[h]
    \centering
    \includegraphics[width=0.9\linewidth]{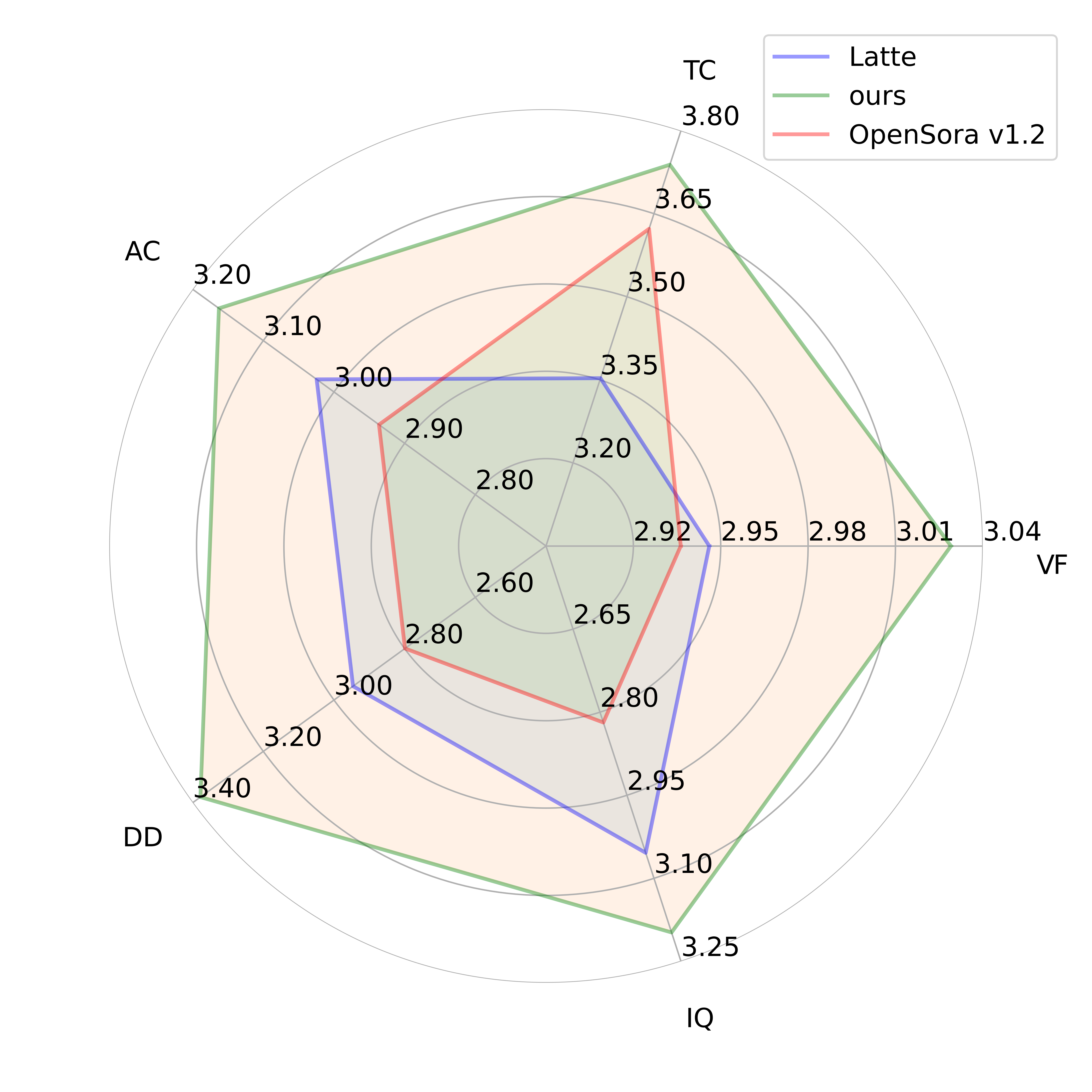} 
    \caption{\textbf{Results of our fine-grained human evaluations.} ``TC'', ``VF'', ``IQ'', ``DD'', ``AC'', are abbreviations for ``text consistency'', ``video faithfulness'', ``image quality'', ``dynamic degree'', and ``action continuity'', respectively.}
    \label{fig:human_evaluation_results}
\end{figure}

Human evaluation plays a vital role in assessing text-to-video models due to its fairness, reliability and interpretability, making it widely used in text-to-video works.
In light of this, we include a fine-grained and comprehensive human evaluation to thoroughly validate the effectiveness of our method.
Specifically, the evaluation is formulated as a set of scoring tasks, where human annotators are required to respectively scoring the delegated one synthesized video considering its performance on video faithfulness, text consistency, action continuity, dynamic degree. 
Each criterion is scored on a scale from 1 to 5, based on clear and discriminative standards.
Since human evaluation is consuming in both time and money, the evaluation is conducted on three competitive text-to-video models, two competitive baseline methods (Latte and OpenSora v1.2), and \ourmethod.
Additionally, all models involved in the human evaluation use a unified generation configuration, with a resolution of 512x512 and a video duration on 2s.
The evaluation results is illustrated in Figure~\ref{fig:human_evaluation_results}.
As can be concluded, our models demonstrate remarkably competitive performance, compared to the baseline methods.
Particularly, our models surpass Latte and OpenSora v1.2  on action continuity, dynamic degree and image quality by a large margin, exhibiting the advantages of \ourmethod.

\begin{table*}[t]
\centering
\caption{\textbf{Results of VBench~\cite{huang2024vbench} evaluation.}
We present a comparative analysis of video generation models evaluated on the VBench benchmark across six key metrics: Subject Consistency, Background Consistency, Image Quality, Human Action, Scene, and Appearance Style. The results highlight the performance of Latte~\cite{ma2024latte}, OpenSora v1.2~\cite{opensora}, and our proposed model (\ourmethod).}
\label{tab:vbench_evaluation}
\begin{tabular}{l|>{\centering\arraybackslash}m{1.5cm}>{\centering\arraybackslash}m{1.5cm}>{\centering\arraybackslash}m{1.5cm}>{\centering\arraybackslash}m{1.5cm}>{\centering\arraybackslash}m{1.5cm}>{\centering\arraybackslash}m{1.5cm}}
\toprule
\multirow{2}{*}{Models} & \multicolumn{1}{c}{Subject} & \multicolumn{1}{c}{Background} & \multicolumn{1}{c}{Image} & \multicolumn{1}{c}{Human} & \multirow{2}{*}{Scene} & \multicolumn{1}{c}{Appearance} \\
& \multicolumn{1}{c}{Consistency} & \multicolumn{1}{c}{Consistency} & \multicolumn{1}{c}{Quality} & \multicolumn{1}{c}{Action} & & \multicolumn{1}{c}{Style} \\
\midrule
Latte~\cite{ma2024latte}                   & 88.9                        & 95.4                            & 61.9                        & 90.0                        & 36.3                     & 23.7 \\
OpenSora v1.2~\cite{opensora}           & 94.5                        & 97.9                            & 60.9                        & 85.8                        & 42.5                     & 23.9 \\
\ourmethod              & 94.7                        & 98.0                            & 62.3                        & 95.0                        & 48.7                     & 24.4 \\
\bottomrule
\end{tabular}
\end{table*}

\begin{figure*}[t]
    \centering
    \includegraphics[width=0.85\linewidth]{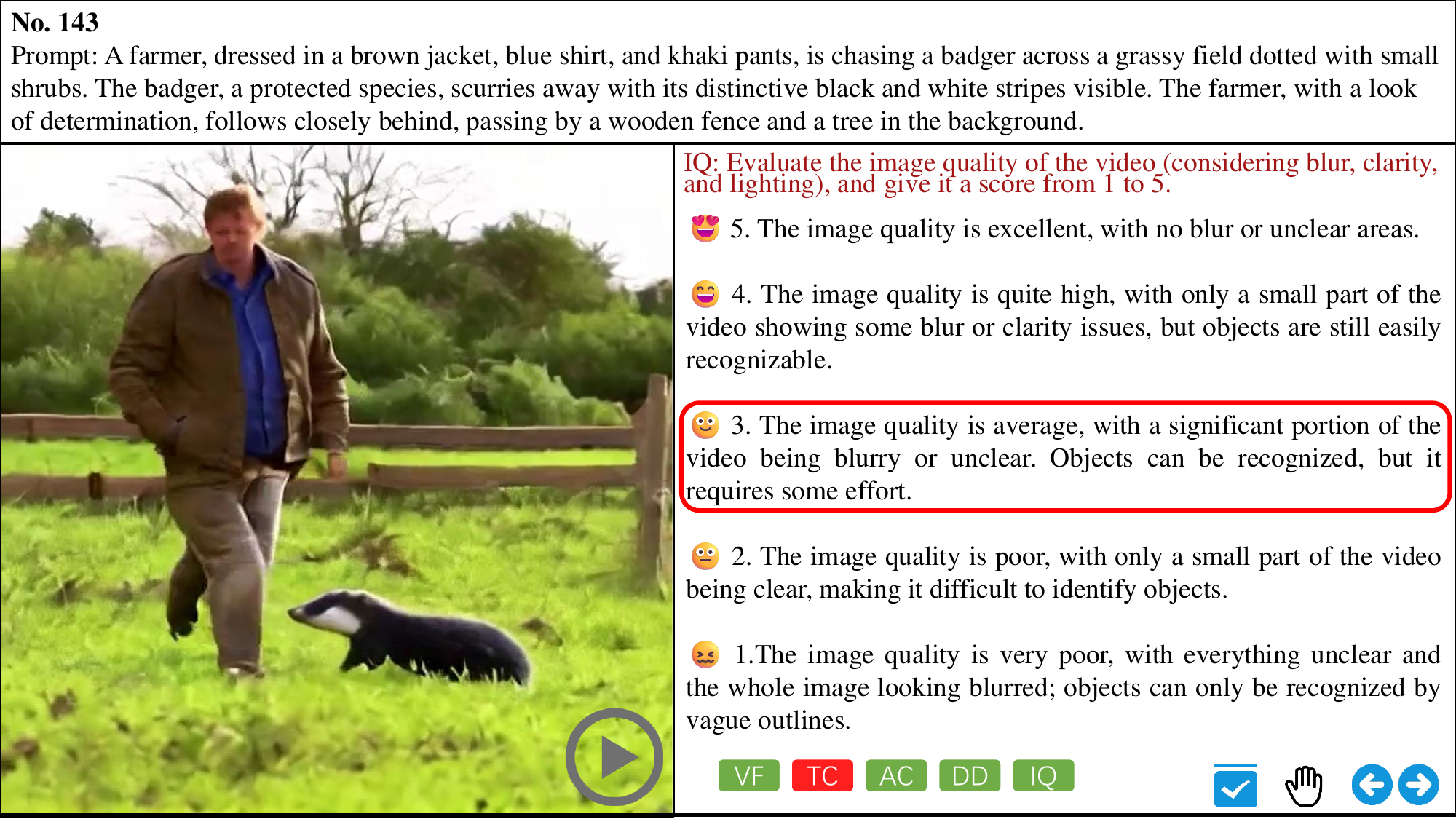} 
    \caption{\textbf{Demonstration of the user interface.} The use interface is specifically designed for our fine-grained human evaluations to lower its costs and enhance its efficiency. Each time, the user interface delegate a synthesized video along with its conditioned prompt to human annotators. Then, the human annotators are supposed to follow the guidance of the user interface and score the videos from five aforementioned perspectives. The labels on the bottom buttons,``TC'', ``VF'', ``IQ'', ``DD'', ``AC'', represent for ``text consistency'', ``video faithfulness'', ``image quality'', ``dynamic degree'', and ``action continuity'', respectively.}
    \label{fig:human_user_interface}
\end{figure*}

\mypara{Human Evaluation Procedure.}
To secure the quality of human evaluation, we employ a series of measures such as prompt selection, user interface design, annotation training and multi-turn quality inspection.
The evaluation is based on model-synthesized videos conditioned on 200 prompts, which are balanced in topic and sourced from various public datasets to guarantee diversity and comprehensiveness.
Besides, throughout of the annotation procedure, the evaluation is conducted via a specifically designed user interface.
As exhibited in Figure~\ref{fig:human_user_interface}, the user interface presents a video alongside its conditioned prompt to human annotators.
Then users are required score the video on a scale from 1 to 5, with each score accompanied by an emoji to clearly convey the extent of their evaluation, along with descriptive sentences about what types of videos correspond to each score. Once a score is selected, it is highlighted by a red column for emphasis.
At the bottom of the user interface are nine buttons. The first five serve as indicators and toggles for the five scoring perspectives; they are initially red and turn green once users complete their scoring tasks. The remaining four buttons facilitate the submission of annotation results, reporting, and navigation between samples.
After the preparation of data and user interface, we train the human annotators about the purpose, formulation, and the principle of the evaluation and annotate several cases as demonstration.
During this training, annotators are encouraged to ask questions about any unclear principles, which we address through further demonstrations and explanations.
Once trained, annotators are free to begin their evaluations. 
Throughout the annotation process, we conduct multi-turn quality inspections, sharing samples that do not meet our standards and providing explanations for their failure.

\subsection{VBench Evaluation}

To further verify the effectiveness of \ourmethod, we also include VBench~\cite{huang2024vbench}, a comprehensive and popular text-to-video benchmark, as a supplement to the FVD and human evaluations.
The results of VBench evaluation are illustrated in Table~\ref{tab:vbench_evaluation}, where our models demonstrate superior performance compared to baseline methods.
Particularly, \ourmethod outperforms baseline methods in terms of ``Human Action'' and ``Scene'' by a large margin, suggesting that advantages of \ourmethod in human-centric and complicated scene video generation.

\subsection{Additional Qualitative Results} 
\begin{figure*}[t]
    \centering
    \includegraphics[width=0.95\textwidth]{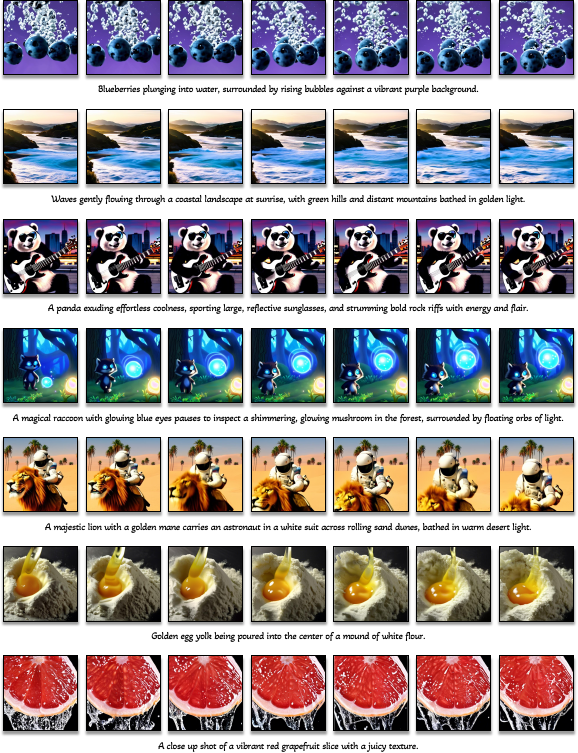} 
    \caption{\textbf{Text-to-video showcases.} The generated videos exhibit significant motion, diverse styles, and demonstrate strong temporal consistency.}
    \label{fig:additional_qual_visualization}
\end{figure*}

More text-to-video examples are shown in Table~\ref{fig:additional_qual_visualization}. We selected prompts from various scenes and styles to generate videos. As a result, \ourmethod more faithfully follows the user’s text instructions in the generated videos. Additionally, \ourmethod demonstrates an understanding not only of what the user specifies in the prompt but also of how these elements exist and interact in the physical world.

\subsection{Bad Case Analysis}
In the analysis of our model’s limitations, we identified several categories of Bad Cases that highlight the challenges in video generation. First, the limited scale of training data significantly impacts the model’s ability to generate videos with diverse semantics and scenes. When encountering less-represented or under-sampled scenarios, the generated videos often fail to capture the desired details or exhibit a lack of diversity. Second, data bias further exacerbates this issue, as overrepresented patterns in the training set lead the model to generate videos that reflect these biases while neglecting less common patterns. This results in outputs that may align poorly with the intended semantics, especially in scenarios requiring fairness or neutrality. Finally, the generation of complex scenes, particularly those involving multiple interacting objects, poses a considerable challenge. The model struggles to maintain coherence in object dynamics and spatial relationships, often producing artifacts or inconsistencies when tasked with such scenarios.

To address these issues, future research can focus on scaling up and diversifying the training data, ensuring better representation across varied semantic and scene distributions. Additionally, techniques such as data augmentation or synthetic data generation could mitigate the effects of data bias. For complex scenes, incorporating stronger spatial-temporal reasoning mechanisms—such as graph-based object modeling or hierarchical architectures—could enhance the model’s ability to capture intricate interactions and maintain coherence across frames. By tackling these challenges, the robustness and generalization capability of video generation models can be significantly improved.

\section{Future work}
While this paper introduces significant advancements in text-to-video generation through a novel dataset curation pipeline and an efficient transformer-based architecture, several avenues remain open for future exploration.

First, although \ourdataset emphasizes high-quality video data with strong text-video alignment, its diversity could be further enhanced by incorporating videos from more diverse domains, cultural contexts, and styles. Expanding the dataset in these directions could improve model generalization and robustness across a wider range of applications. 

Second, while \ourmethodbold demonstrates computational efficiency through a progressive four-stage training strategy, future work could focus on optimizing resource utilization further. For example, leveraging sparse attention mechanisms or integrating adaptive computation methods may enable scaling to larger models or datasets without proportional increases in computational costs.

Third, our current architecture prioritizes spatial-temporal decoupling for effective video generation, but future research could explore joint spatial-temporal modeling or hybrid approaches that balance efficiency and coherence. Additionally, incorporating audio-visual alignment into the framework could unlock possibilities for multimodal video generation, further enriching the user experience.

Finally, as text-to-video models become more accessible, addressing ethical concerns such as content bias, misuse, and fairness remains critical. Future work could focus on developing robust content filtering mechanisms, improving interpretability, and aligning generation with ethical guidelines and societal values.

We envision these future directions not only enhancing the technical capabilities of text-to-video generation but also promoting responsible and impactful applications.

%% file: main.bbl
\begin{thebibliography}{51}
\providecommand{\natexlab}[1]{#1}
\providecommand{\url}[1]{\texttt{#1}}
\expandafter\ifx\csname urlstyle\endcsname\relax
  \providecommand{\doi}[1]{doi: #1}\else
  \providecommand{\doi}{doi: \begingroup \urlstyle{rm}\Url}\fi

\bibitem[Anne~Hendricks et~al.(2017)Anne~Hendricks, Wang, Shechtman, Sivic, Darrell, and Russell]{anne2017ldidemo}
Lisa Anne~Hendricks, Oliver Wang, Eli Shechtman, Josef Sivic, Trevor Darrell, and Bryan Russell.
\newblock Localizing moments in video with natural language.
\newblock In \emph{Proceedings of the IEEE international conference on computer vision}, pages 5803--5812, 2017.

\bibitem[Blattmann et~al.(2023{\natexlab{a}})Blattmann, Dockhorn, Kulal, Mendelevitch, Kilian, Lorenz, Levi, English, Voleti, Letts, et~al.]{blattmann2023svd}
Andreas Blattmann, Tim Dockhorn, Sumith Kulal, Daniel Mendelevitch, Maciej Kilian, Dominik Lorenz, Yam Levi, Zion English, Vikram Voleti, Adam Letts, et~al.
\newblock Stable video diffusion: Scaling latent video diffusion models to large datasets.
\newblock \emph{arXiv preprint arXiv:2311.15127}, 2023{\natexlab{a}}.

\bibitem[Blattmann et~al.(2023{\natexlab{b}})Blattmann, Rombach, Ling, Dockhorn, Kim, Fidler, and Kreis]{blattmann2023align}
Andreas Blattmann, Robin Rombach, Huan Ling, Tim Dockhorn, Seung~Wook Kim, Sanja Fidler, and Karsten Kreis.
\newblock Align your latents: High-resolution video synthesis with latent diffusion models.
\newblock In \emph{Proceedings of the IEEE/CVF Conference on Computer Vision and Pattern Recognition}, pages 22563--22575, 2023{\natexlab{b}}.

\bibitem[Blattmann et~al.(2023{\natexlab{c}})Blattmann, Rombach, Ling, Dockhorn, Kim, Fidler, and Kreis]{blattmann2023videoldm}
Andreas Blattmann, Robin Rombach, Huan Ling, Tim Dockhorn, Seung~Wook Kim, Sanja Fidler, and Karsten Kreis.
\newblock Align your latents: High-resolution video synthesis with latent diffusion models.
\newblock In \emph{Proceedings of the IEEE/CVF Conference on Computer Vision and Pattern Recognition}, pages 22563--22575, 2023{\natexlab{c}}.

\bibitem[Brooks et~al.(2024)Brooks, Peebles, Holmes, DePue, Guo, Jing, Schnurr, Taylor, Luhman, Luhman, et~al.]{brooks2024video}
Tim Brooks, Bill Peebles, Connor Holmes, Will DePue, Yufei Guo, Li Jing, David Schnurr, Joe Taylor, Troy Luhman, Eric Luhman, et~al.
\newblock Video generation models as world simulators. 2024.
\newblock \emph{URL https://openai. com/research/video-generation-models-as-world-simulators}, 3, 2024.

\bibitem[Caba~Heilbron et~al.(2015)Caba~Heilbron, Escorcia, Ghanem, and Carlos~Niebles]{caba2015activitynet}
Fabian Caba~Heilbron, Victor Escorcia, Bernard Ghanem, and Juan Carlos~Niebles.
\newblock Activitynet: A large-scale video benchmark for human activity understanding.
\newblock In \emph{Proceedings of the ieee conference on computer vision and pattern recognition}, pages 961--970, 2015.

\bibitem[Castellano(2014)]{pyscenedetect}
Brandon Castellano.
\newblock Pyscenedetect.
\newblock \url{https://github.com/Breakthrough/PySceneDetect}, 2014.

\bibitem[Chen and Dolan(2011)]{chen2011msvd}
David Chen and William~B Dolan.
\newblock Collecting highly parallel data for paraphrase evaluation.
\newblock In \emph{Proceedings of the 49th annual meeting of the association for computational linguistics: human language technologies}, pages 190--200, 2011.

\bibitem[Chen et~al.()Chen, Jincheng, Chongjian, Yao, Xie, Wang, Kwok, Luo, Lu, and Li]{chenpixart}
Junsong Chen, YU Jincheng, GE Chongjian, Lewei Yao, Enze Xie, Zhongdao Wang, James Kwok, Ping Luo, Huchuan Lu, and Zhenguo Li.
\newblock Pixart-alpha: Fast training of diffusion transformer for photorealistic text-to-image synthesis.
\newblock In \emph{The Twelfth International Conference on Learning Representations}.

\bibitem[Chen et~al.(2024)Chen, Siarohin, Menapace, Deyneka, Chao, Jeon, Fang, Lee, Ren, Yang, et~al.]{chen2024panda}
Tsai-Shien Chen, Aliaksandr Siarohin, Willi Menapace, Ekaterina Deyneka, Hsiang-wei Chao, Byung~Eun Jeon, Yuwei Fang, Hsin-Ying Lee, Jian Ren, Ming-Hsuan Yang, et~al.
\newblock Panda-70m: Captioning 70m videos with multiple cross-modality teachers.
\newblock In \emph{Proceedings of the IEEE/CVF Conference on Computer Vision and Pattern Recognition}, pages 13320--13331, 2024.

\bibitem[Chen et~al.(2023)Chen, Wang, Zhang, Zhuang, Ma, Yu, Wang, Lin, Qiao, and Liu]{chen2023seine}
Xinyuan Chen, Yaohui Wang, Lingjun Zhang, Shaobin Zhuang, Xin Ma, Jiashuo Yu, Yali Wang, Dahua Lin, Yu Qiao, and Ziwei Liu.
\newblock Seine: Short-to-long video diffusion model for generative transition and prediction.
\newblock In \emph{The Twelfth International Conference on Learning Representations}, 2023.

\bibitem[Dubey et~al.(2024)Dubey, Jauhri, Pandey, Kadian, Al-Dahle, Letman, Mathur, Schelten, Yang, Fan, et~al.]{dubey2024llama}
Abhimanyu Dubey, Abhinav Jauhri, Abhinav Pandey, Abhishek Kadian, Ahmad Al-Dahle, Aiesha Letman, Akhil Mathur, Alan Schelten, Amy Yang, Angela Fan, et~al.
\newblock The llama 3 herd of models.
\newblock \emph{arXiv preprint arXiv:2407.21783}, 2024.

\bibitem[Esser et~al.(2021)Esser, Rombach, and Ommer]{esser2021vqgan}
Patrick Esser, Robin Rombach, and Bjorn Ommer.
\newblock Taming transformers for high-resolution image synthesis.
\newblock In \emph{Proceedings of the IEEE/CVF conference on computer vision and pattern recognition}, pages 12873--12883, 2021.

\bibitem[Ge et~al.(2023)Ge, Nah, Liu, Poon, Tao, Catanzaro, Jacobs, Huang, Liu, and Balaji]{ge2023pyoco}
Songwei Ge, Seungjun Nah, Guilin Liu, Tyler Poon, Andrew Tao, Bryan Catanzaro, David Jacobs, Jia-Bin Huang, Ming-Yu Liu, and Yogesh Balaji.
\newblock Preserve your own correlation: A noise prior for video diffusion models.
\newblock In \emph{Proceedings of the IEEE/CVF International Conference on Computer Vision}, pages 22930--22941, 2023.

\bibitem[Guo et~al.(2023)Guo, Yang, Rao, Liang, Wang, Qiao, Agrawala, Lin, and Dai]{guo2023animatediff}
Yuwei Guo, Ceyuan Yang, Anyi Rao, Zhengyang Liang, Yaohui Wang, Yu Qiao, Maneesh Agrawala, Dahua Lin, and Bo Dai.
\newblock Animatediff: Animate your personalized text-to-image diffusion models without specific tuning.
\newblock \emph{arXiv preprint arXiv:2307.04725}, 2023.

\bibitem[Gupta et~al.(2023)Gupta, Yu, Sohn, Gu, Hahn, Fei-Fei, Essa, Jiang, and Lezama]{gupta2023walt}
Agrim Gupta, Lijun Yu, Kihyuk Sohn, Xiuye Gu, Meera Hahn, Li Fei-Fei, Irfan Essa, Lu Jiang, and Jos{\'e} Lezama.
\newblock Photorealistic video generation with diffusion models.
\newblock \emph{arXiv preprint arXiv:2312.06662}, 2023.

\bibitem[He et~al.(2022)He, Yang, Zhang, Shan, and Chen]{he2022lvdm}
Yingqing He, Tianyu Yang, Yong Zhang, Ying Shan, and Qifeng Chen.
\newblock Latent video diffusion models for high-fidelity long video generation.
\newblock \emph{arXiv preprint arXiv:2211.13221}, 2022.

\bibitem[Ho et~al.(2022{\natexlab{a}})Ho, Chan, Saharia, Whang, Gao, Gritsenko, Kingma, Poole, Norouzi, Fleet, et~al.]{ho2022imagen}
Jonathan Ho, William Chan, Chitwan Saharia, Jay Whang, Ruiqi Gao, Alexey Gritsenko, Diederik~P Kingma, Ben Poole, Mohammad Norouzi, David~J Fleet, et~al.
\newblock Imagen video: High definition video generation with diffusion models.
\newblock \emph{arXiv preprint arXiv:2210.02303}, 2022{\natexlab{a}}.

\bibitem[Ho et~al.(2022{\natexlab{b}})Ho, Salimans, Gritsenko, Chan, Norouzi, and Fleet]{ho2022video}
Jonathan Ho, Tim Salimans, Alexey Gritsenko, William Chan, Mohammad Norouzi, and David~J Fleet.
\newblock Video diffusion models.
\newblock \emph{Advances in Neural Information Processing Systems}, 35:\penalty0 8633--8646, 2022{\natexlab{b}}.

\bibitem[Hong et~al.(2022)Hong, Ding, Zheng, Liu, and Tang]{hong2022cogvideo}
Wenyi Hong, Ming Ding, Wendi Zheng, Xinghan Liu, and Jie Tang.
\newblock Cogvideo: Large-scale pretraining for text-to-video generation via transformers.
\newblock \emph{arXiv preprint arXiv:2205.15868}, 2022.

\bibitem[Huang et~al.(2024)Huang, He, Yu, Zhang, Si, Jiang, Zhang, Wu, Jin, Chanpaisit, et~al.]{huang2024vbench}
Ziqi Huang, Yinan He, Jiashuo Yu, Fan Zhang, Chenyang Si, Yuming Jiang, Yuanhan Zhang, Tianxing Wu, Qingyang Jin, Nattapol Chanpaisit, et~al.
\newblock Vbench: Comprehensive benchmark suite for video generative models.
\newblock In \emph{Proceedings of the IEEE/CVF Conference on Computer Vision and Pattern Recognition}, pages 21807--21818, 2024.

\bibitem[Jiang et~al.(2021)Jiang, Chang, and Wang]{jiang2021transgan}
Yifan Jiang, Shiyu Chang, and Zhangyang Wang.
\newblock Transgan: Two pure transformers can make one strong gan, and that can scale up.
\newblock \emph{Advances in Neural Information Processing Systems}, 34:\penalty0 14745--14758, 2021.

\bibitem[Lee et~al.(2021)Lee, Chang, Jiang, Zhang, Tu, and Liu]{lee2021vitgan}
Kwonjoon Lee, Huiwen Chang, Lu Jiang, Han Zhang, Zhuowen Tu, and Ce Liu.
\newblock Vitgan: Training gans with vision transformers.
\newblock \emph{arXiv preprint arXiv:2107.04589}, 2021.

\bibitem[Li et~al.(2022)Li, Liu, Guo, Yin, Jiang, Du, Du, Zhu, Lai, Hu, et~al.]{li2022pp_ocr}
Chenxia Li, Weiwei Liu, Ruoyu Guo, Xiaoting Yin, Kaitao Jiang, Yongkun Du, Yuning Du, Lingfeng Zhu, Baohua Lai, Xiaoguang Hu, et~al.
\newblock Pp-ocrv3: More attempts for the improvement of ultra lightweight ocr system.
\newblock \emph{arXiv preprint arXiv:2206.03001}, 2022.

\bibitem[Lin et~al.(2024)Lin, Yin, Ping, Molchanov, Shoeybi, and Han]{lin2024vila}
Ji Lin, Hongxu Yin, Wei Ping, Pavlo Molchanov, Mohammad Shoeybi, and Song Han.
\newblock Vila: On pre-training for visual language models.
\newblock In \emph{Proceedings of the IEEE/CVF Conference on Computer Vision and Pattern Recognition}, pages 26689--26699, 2024.

\bibitem[Lu et~al.(2023)Lu, Yang, Fei, Huo, Lu, Luo, and Ding]{lu2023vdt}
Haoyu Lu, Guoxing Yang, Nanyi Fei, Yuqi Huo, Zhiwu Lu, Ping Luo, and Mingyu Ding.
\newblock Vdt: General-purpose video diffusion transformers via mask modeling.
\newblock \emph{arXiv preprint arXiv:2305.13311}, 2023.

\bibitem[Ma et~al.(2024)Ma, Wang, Jia, Chen, Liu, Li, Chen, and Qiao]{ma2024latte}
Xin Ma, Yaohui Wang, Gengyun Jia, Xinyuan Chen, Ziwei Liu, Yuan-Fang Li, Cunjian Chen, and Yu Qiao.
\newblock Latte: Latent diffusion transformer for video generation.
\newblock \emph{arXiv preprint arXiv:2401.03048}, 2024.

\bibitem[Miech et~al.(2019)Miech, Zhukov, Alayrac, Tapaswi, Laptev, and Sivic]{miech2019howto100m}
Antoine Miech, Dimitri Zhukov, Jean-Baptiste Alayrac, Makarand Tapaswi, Ivan Laptev, and Josef Sivic.
\newblock Howto100m: Learning a text-video embedding by watching hundred million narrated video clips.
\newblock In \emph{Proceedings of the IEEE/CVF international conference on computer vision}, pages 2630--2640, 2019.

\bibitem[Peebles and Xie(2023)]{peebles2023dit}
William Peebles and Saining Xie.
\newblock Scalable diffusion models with transformers.
\newblock In \emph{Proceedings of the IEEE/CVF International Conference on Computer Vision}, pages 4195--4205, 2023.

\bibitem[Radford et~al.(2021)Radford, Kim, Hallacy, Ramesh, Goh, Agarwal, Sastry, Askell, Mishkin, Clark, et~al.]{radford2021clip}
Alec Radford, Jong~Wook Kim, Chris Hallacy, Aditya Ramesh, Gabriel Goh, Sandhini Agarwal, Girish Sastry, Amanda Askell, Pamela Mishkin, Jack Clark, et~al.
\newblock Learning transferable visual models from natural language supervision.
\newblock In \emph{International conference on machine learning}, pages 8748--8763. PMLR, 2021.

\bibitem[Rohrbach et~al.(2015)Rohrbach, Rohrbach, Tandon, and Schiele]{rohrbach2015lsmdc}
Anna Rohrbach, Marcus Rohrbach, Niket Tandon, and Bernt Schiele.
\newblock A dataset for movie description.
\newblock In \emph{Proceedings of the IEEE conference on computer vision and pattern recognition}, pages 3202--3212, 2015.

\bibitem[Rombach et~al.(2022)Rombach, Blattmann, Lorenz, Esser, and Ommer]{rombach2022high}
Robin Rombach, Andreas Blattmann, Dominik Lorenz, Patrick Esser, and Bj{\"o}rn Ommer.
\newblock High-resolution image synthesis with latent diffusion models.
\newblock In \emph{Proceedings of the IEEE/CVF conference on computer vision and pattern recognition}, pages 10684--10695, 2022.

\bibitem[Saharia et~al.(2022)Saharia, Chan, Saxena, Li, Whang, Denton, Ghasemipour, Gontijo~Lopes, Karagol~Ayan, Salimans, et~al.]{saharia2022photorealistic}
Chitwan Saharia, William Chan, Saurabh Saxena, Lala Li, Jay Whang, Emily~L Denton, Kamyar Ghasemipour, Raphael Gontijo~Lopes, Burcu Karagol~Ayan, Tim Salimans, et~al.
\newblock Photorealistic text-to-image diffusion models with deep language understanding.
\newblock \emph{Advances in neural information processing systems}, 35:\penalty0 36479--36494, 2022.

\bibitem[Schuhmann et~al.(2022)Schuhmann, Beaumont, Vencu, Gordon, Wightman, Cherti, Coombes, Katta, Mullis, Wortsman, et~al.]{schuhmann2022laion}
Christoph Schuhmann, Romain Beaumont, Richard Vencu, Cade Gordon, Ross Wightman, Mehdi Cherti, Theo Coombes, Aarush Katta, Clayton Mullis, Mitchell Wortsman, et~al.
\newblock Laion-5b: An open large-scale dataset for training next generation image-text models.
\newblock \emph{Advances in Neural Information Processing Systems}, 35:\penalty0 25278--25294, 2022.

\bibitem[Singer et~al.(2022)Singer, Polyak, Hayes, Yin, An, Zhang, Hu, Yang, Ashual, Gafni, et~al.]{singer2022make-a-video}
Uriel Singer, Adam Polyak, Thomas Hayes, Xi Yin, Jie An, Songyang Zhang, Qiyuan Hu, Harry Yang, Oron Ashual, Oran Gafni, et~al.
\newblock Make-a-video: Text-to-video generation without text-video data.
\newblock \emph{arXiv preprint arXiv:2209.14792}, 2022.

\bibitem[Skorokhodov et~al.(2022)Skorokhodov, Tulyakov, and Elhoseiny]{skorokhodov2022stylegan}
Ivan Skorokhodov, Sergey Tulyakov, and Mohamed Elhoseiny.
\newblock Stylegan-v: A continuous video generator with the price, image quality and perks of stylegan2.
\newblock In \emph{Proceedings of the IEEE/CVF conference on computer vision and pattern recognition}, pages 3626--3636, 2022.

\bibitem[Soomro et~al.(2012)Soomro, Zamir, and Shah]{soomro2012ucf101}
Khurram Soomro, Amir~Roshan Zamir, and Mubarak Shah.
\newblock Ucf101: A dataset of 101 human actions classes from videos in the wild.
\newblock \emph{arXiv preprint arXiv:1212.0402}, 2012.

\bibitem[Tan et~al.(2025)Tan, Yang, Qin, Yang, Qian, Zhou, Zhang, and Li]{tan2025empirical}
Zhiyu Tan, Mengping Yang, Luozheng Qin, Hao Yang, Ye Qian, Qiang Zhou, Cheng Zhang, and Hao Li.
\newblock An empirical study and analysis of text-to-image generation using large language model-powered textual representation.
\newblock In \emph{European Conference on Computer Vision}, pages 472--489. Springer, 2025.

\bibitem[Teed and Deng(2020)]{teed2020raft}
Zachary Teed and Jia Deng.
\newblock Raft: Recurrent all-pairs field transforms for optical flow.
\newblock In \emph{Computer Vision--ECCV 2020: 16th European Conference, Glasgow, UK, August 23--28, 2020, Proceedings, Part II 16}, pages 402--419. Springer, 2020.

\bibitem[Unterthiner et~al.(2018)Unterthiner, Van~Steenkiste, Kurach, Marinier, Michalski, and Gelly]{unterthiner2018towards_fvd}
Thomas Unterthiner, Sjoerd Van~Steenkiste, Karol Kurach, Raphael Marinier, Marcin Michalski, and Sylvain Gelly.
\newblock Towards accurate generative models of video: A new metric \& challenges.
\newblock \emph{arXiv preprint arXiv:1812.01717}, 2018.

\bibitem[Vaswani(2017)]{vaswani2017attention}
A Vaswani.
\newblock Attention is all you need.
\newblock \emph{Advances in Neural Information Processing Systems}, 2017.

\bibitem[Wang et~al.(2023)Wang, Yuan, Chen, Zhang, Wang, and Zhang]{wang2023modelscope}
Jiuniu Wang, Hangjie Yuan, Dayou Chen, Yingya Zhang, Xiang Wang, and Shiwei Zhang.
\newblock Modelscope text-to-video technical report.
\newblock \emph{arXiv preprint arXiv:2308.06571}, 2023.

\bibitem[Wang et~al.(2019)Wang, Wu, Chen, Li, Wang, and Wang]{wang2019vatex}
Xin Wang, Jiawei Wu, Junkun Chen, Lei Li, Yuan-Fang Wang, and William~Yang Wang.
\newblock Vatex: A large-scale, high-quality multilingual dataset for video-and-language research.
\newblock In \emph{Proceedings of the IEEE/CVF international conference on computer vision}, pages 4581--4591, 2019.

\bibitem[Wang et~al.(2024)Wang, Ma, Chen, Chen, Dantcheva, Dai, and Qiao]{wang2024leo}
Yaohui Wang, Xin Ma, Xinyuan Chen, Cunjian Chen, Antitza Dantcheva, Bo Dai, and Yu Qiao.
\newblock Leo: Generative latent image animator for human video synthesis.
\newblock \emph{International Journal of Computer Vision}, pages 1--13, 2024.

\bibitem[Xu et~al.(2016)Xu, Mei, Yao, and Rui]{xu2016msrvtt}
Jun Xu, Tao Mei, Ting Yao, and Yong Rui.
\newblock Msr-vtt: A large video description dataset for bridging video and language.
\newblock In \emph{Proceedings of the IEEE conference on computer vision and pattern recognition}, pages 5288--5296, 2016.

\bibitem[Xue et~al.(2022)Xue, Hang, Zeng, Sun, Liu, Yang, Fu, and Guo]{xue2022hdvila}
Hongwei Xue, Tiankai Hang, Yanhong Zeng, Yuchong Sun, Bei Liu, Huan Yang, Jianlong Fu, and Baining Guo.
\newblock Advancing high-resolution video-language representation with large-scale video transcriptions.
\newblock In \emph{Proceedings of the IEEE/CVF Conference on Computer Vision and Pattern Recognition}, pages 5036--5045, 2022.

\bibitem[Yan et~al.(2021)Yan, Zhang, Abbeel, and Srinivas]{yan2021videogpt}
Wilson Yan, Yunzhi Zhang, Pieter Abbeel, and Aravind Srinivas.
\newblock Videogpt: Video generation using vq-vae and transformers.
\newblock \emph{arXiv preprint arXiv:2104.10157}, 2021.

\bibitem[Zellers et~al.(2021)Zellers, Lu, Hessel, Yu, Park, Cao, Farhadi, and Choi]{zellers2021yt180m}
Rowan Zellers, Ximing Lu, Jack Hessel, Youngjae Yu, Jae~Sung Park, Jize Cao, Ali Farhadi, and Yejin Choi.
\newblock Merlot: Multimodal neural script knowledge models.
\newblock \emph{Advances in neural information processing systems}, 34:\penalty0 23634--23651, 2021.

\bibitem[Zheng et~al.(2024)Zheng, Peng, Yang, Shen, Li, Liu, Zhou, Li, and You]{opensora}
Zangwei Zheng, Xiangyu Peng, Tianji Yang, Chenhui Shen, Shenggui Li, Hongxin Liu, Yukun Zhou, Tianyi Li, and Yang You.
\newblock Open-sora: Democratizing efficient video production for all, 2024.

\bibitem[Zhou et~al.(2022)Zhou, Wang, Yan, Lv, Zhu, and Feng]{zhou2022magicvideo}
Daquan Zhou, Weimin Wang, Hanshu Yan, Weiwei Lv, Yizhe Zhu, and Jiashi Feng.
\newblock Magicvideo: Efficient video generation with latent diffusion models.
\newblock \emph{arXiv preprint arXiv:2211.11018}, 2022.

\bibitem[Zhou et~al.(2018)Zhou, Xu, and Corso]{zhou2018youcook2}
Luowei Zhou, Chenliang Xu, and Jason Corso.
\newblock Towards automatic learning of procedures from web instructional videos.
\newblock In \emph{Proceedings of the AAAI Conference on Artificial Intelligence}, 2018.

\end{thebibliography}
